\documentclass{article} 
\usepackage{iclr2026_conference,times}

\usepackage{amsmath,amsfonts,bm}

\def\eqref#1{equation~\ref{#1}}

\def\1{\bm{1}}

\DeclareMathAlphabet{\mathsfit}{\encodingdefault}{\sfdefault}{m}{sl}
\SetMathAlphabet{\mathsfit}{bold}{\encodingdefault}{\sfdefault}{bx}{n}

\usepackage{hyperref}
\usepackage{url}

\usepackage{graphicx}
\usepackage{subcaption}
\usepackage{algorithm}
\usepackage{amsmath,amssymb}
\usepackage{dsfont}
\usepackage{algpseudocode}

\usepackage{wrapfig}
\usepackage{enumitem}
\usepackage{multirow}
\usepackage{tabularx}
\title{Evaluation of Multi-Turn Consistency in LLM Agents: Survival Analysis and Failure-Rationale Taxonomy}

\author{Igor Bogdanov \\
Systems and Computer Engineering \\
Carleton University \\
Ottawa, ON, Canada \\
\texttt{igorbogdanov@cmail.carleton.ca} \\
\And 
Olga Manakina \\
Department of Cognitive Science\\
Carleton University\\
Ottawa, ON, Canada \\
\texttt{olgamanakina@cmail.carleton.ca} \\
\And
Chung-Horng Lung \\
Systems and Computer Engineering \\
Carleton University \\
Ottawa, ON, Canada \\
\texttt{chlung@sce.carleton.ca} \\
}

\iclrfinalcopy 
\begin{document}

\maketitle
\lhead{Accepted at the ICLR 2026 Workshop on Logical Reasoning of Large Language Models}

\begin{abstract}

   Large language model (LLM) agents may perform well on isolated tasks yet drift into inconsistency over extended interaction. We evaluate temporal consistency in a controlled 20-step multi-agent setting inspired by delayed-gratification studies. At each step, an agent chooses between continuing to delay a reward or claiming it immediately (terminating the episode). Across a full-factorial manipulation of social visibility (private vs public), persona stressors, and deliberation policy, we run 84,540 trajectories spanning 8 model families. Treating the first reward-claim as a time-to-event outcome, we estimate Kaplan-Meier survival curves and fit discrete-time hazard regression 
   to quantify how experimental factors shift failure risk over time. Then, to analyze rationales and language patterns associated with failure, we build a seven-category taxonomy from 13,780 deliberation traces from agents who choose to terminate the episode, using an LLM-assisted labeling paired with human audit ($\kappa=0.83$). Rationale profiles change systematically with time and context: early failures are more impulse-driven, later failures more fatigue- and cost–benefit-framed, while public settings increase norm-oriented justifications. We also find a deliberation-inconsistency association: among failures, longer deliberation correlates with higher rates of intra-rationale contradiction (simultaneous pro-delay and pro-claim statements), challenging the assumption that more reasoning text implies greater consistency. Together, the survival and rationale analyses reveal distinct temporal reliability regimes and model-specific "failure fingerprints", offering an evaluation lens for diagnosing inconsistency in multi-turn agent behavior.

\end{abstract}

\section{Introduction}

LLM agents can demonstrate coherence on single-turn tasks but become inconsistent in multi-turn interactions, drifting in commitments and occasionally producing contradictions that undermine reliability. Existing agent benchmarks typically report terminal success rates \citep{liu2023agentbench,wang2024mint,ma2024agentboard,zhang2024twentyq}, but provide limited insight into how inconsistency emerges over time and what decision narratives precede constraint violations. When an agent abandons an instruction or violates a rule, what rationale does it produce? Do these rationales and their internal consistency vary systematically across time, social context, and model family? Answering these questions requires both a controlled environment that induces failures under known conditions and a method for semantically characterizing the deliberation traces that accompany them.

We address this gap by introducing a semantic taxonomy of pre-failure rationales together with a time-resolved evaluation suite based on time-to-failure (time-to-event) tracking. We adapt the classic delayed-gratification paradigm, known as the Stanford marshmallow test \citep{mischel1972marshmallow} into a discrete-time commitment task: at each of 20 steps, an agent must choose between claiming an immediate reward (terminating the episode) or continuing to defer for a larger payoff. This minimal binary decision isolates temporal drift, sustained instruction-following under pressure, and social cascades \citep{du2023improving,wu2024autogen,laban2025lost}, while generating rich natural-language deliberation traces for analysis. We treat contradiction within these traces as a minimal, model-agnostic signal of logical inconsistency in multi-turn decision-making.

\paragraph{Contributions.}
Our contribution is an evaluation methodology (factorial perturbations + time-to-failure modeling) paired with failure-rationale cartography instantiated in a controlled setting. Specifically, we present:

\textit{Rationale-based diagnostics of long-horizon inconsistency.} We introduce a semantic taxonomy of pre-failure traces and a linguistically grounded feature set (modal/temporal/hedonic and discourse-structural markers, commitment/uncertainty cues, and an intra-rationale inconsistency signal) to map how rationales vary across models, time, and experimental conditions.

\textit{A controlled multi-turn temporal consistency benchmark.} We operationalize delayed reward as a 20-step claim-or-defer setting and evaluate 8 model families under full-factorial manipulations (social visibility, persona stressors, and self-questioning policy), yielding 84,540 trajectories and 13,780 labeled pre-failure traces.

\textit{Time-to-event reliability via survival/hazard modeling.} We treat failure as a time-to-event process and estimate the effect of experimental factors on survival and hazard over the interaction horizon, enabling comparisons beyond terminal success rates and linking temporal reliability patterns to shifts in pre-failure rationales.

\paragraph{Threats to validity.}
Pre-failure traces are self-reported justifications, so we analyze them as behavioral text artifacts and emphasize stable associations across controlled conditions and time. Our focus is temporal consistency and contradiction signals in multi-turn settings, not formal logical validity proofs. Taxonomy boundaries and prompt format can influence labels and features. To address these concerns, we ground our taxonomy in a human-developed codebook (N=100), verify LLM labels against human audit ($\kappa=0.83$), and hold prompts and instrumentation fixed across all conditions. We acknowledge that this controlled setting is not a proxy for full deployment.

\section{Related Work}

\textbf{Failure and inconsistency taxonomies for LLM reasoning.}\\
Recent work argues that evaluation should characterize \emph{what kind} of failure occurs, not just how often. Hallucination taxonomies \citep{huang2023hallucination} are operationalized in suites like HalluLens \citep{bang-etal-2025-hallulens} and probed via consistency-based detectors \citep{manakul-etal-2023-selfcheckgpt}. Structured reasoning-error taxonomies further analyze where self-verification breaks down \citep{hong-etal-2024-closer}. However, these approaches are typically studied in single-turn settings, leaving open how inconsistency manifests and evolves over extended interactions where prior decisions and stated commitments accumulate. We extend this line by deriving a taxonomy from multi-turn agent deliberations and analyzing how expressed failure narratives—and their internal consistency—shift with time and experimental conditions.

\textbf{Reasoning traces, self-verification, and faithfulness.}\\
Chain-of-thought and other reasoning traces raise faithfulness concerns: perturbing traces can change answers, and faithfulness varies across tasks and scales \citep{lanham2023cotfaithfulness,tutek2025measuring}. Recent surveys propose organizing trace evaluation along dimensions such as validity, coherence, and utility \citep{lee-hockenmaier-2025-evaluating}. We therefore treat pre-failure traces as behavioral text artifacts: diagnostics of expressed reasons that are anchored to objective time-to-failure events in the interaction. This stance supports interpretability without overclaiming access to internal causality, while still enabling principled signals of inconsistency (e.g., intra-trace self-contradiction). We observe an association between more elaborate deliberation and higher rates of contradiction in our setting, which speaks to debates about when traces should be trusted.

\textbf{Long-horizon agent evaluation.}\\
Agent benchmarks evaluate multi-turn decision-making and tool use \citep{liu2023agentbench,wang2024mint,ma2024agentboard}, and simulation studies show that errors can compound over turns \citep{laban2025lost}. Broader evaluation frameworks advocate multi-metric reporting \citep{liang2023helm}. Yet many existing evaluations emphasize terminal outcomes, which can obscure when failures emerge and whether there are distinct temporal regimes of reliability. We complement terminal success metrics with time-to-failure analysis (survival curves, hazard models) to separate early- versus late-stage failure risk and to connect temporal reliability patterns to shifts in failure rationales.

\textbf{LLMs as behavioral subjects and controlled testbeds.}\\
A growing literature treats LLMs as experimental subjects using established behavioral tasks \citep{hagendorff2023thinkingfastslow,strachan2024tom,ross2024llmeconomicus,sartori2023psychsciences,li2025llmscogsci}. We draw on delayed-gratification paradigms \citep{mischel1972marshmallow,metcalfe1999hot,kidd2013rational,watts2018revisiting,casey2011behavioral} not as a psychological claim about human cognition, but as a controlled environment that reliably induces temporally extended commitment pressure and produces rich deliberation traces. In this sense, the task functions as a consistency stress test for multi-turn agent behavior, enabling systematic study of how reasoning narratives and inconsistency signals change across time and interaction context.

\section{Methodology}

\subsection{Temporal Consistency Task (Claim-or-Defer)}
We study temporal consistency in a controlled, finite-horizon, discrete-time decision task inspired by delayed-reward paradigms (background in Appendix~\ref{app:marshmallow_background}). Each episode lasts for $T{=}20$ steps (indexed 0-19). At each step $t\in\{1,\dots,T\}$, the agent chooses between (i) \textsc{Defer} (continue the episode toward a larger delayed payoff) and (ii) \textsc{Claim} (take a smaller immediate payoff and terminate the episode). In implementation, actions are emitted as constrained strings $\{\text{"I wait"},\text{"I eat the marshmallow"}\}$, which we refer to as \textsc{Defer} and \textsc{Claim}, respectively. Claiming yields an immediate reward (+1) and ends the episode; deferring advances to the next step. Agents that reach $T$ without claiming receive the delayed payoff (+2).

Observations include the current step index and, in broadcast (social) conditions, a summary of recent peer actions. Agents may optionally engage in internal deliberation via the \texttt{raise\_a\_question} tool, subject to a per-step budget $C$; tool calls do not alter the environment state. Full environment formalization is provided in Appendix~\ref{app:pomdp}.

\subsection{Experimental Design (Full-Factorial Perturbations)}
We run a full-factorial design crossing three families of perturbations intended to stress temporal consistency:
(i) \textbf{social visibility} (isolated vs.\ broadcast, where agents observe peers' actions),
(ii) \textbf{persona stressors} (age: child/adult/senior/none; hedonic drive: crave/like/neutral/none),
and (iii) \textbf{deliberation policy} (self-query tool required vs.\ optional).
This yields 64 condition combinations per model family. Prompts and instrumentation are held fixed across conditions, with only the targeted factors varying.

\subsection{Models, Runs, and Data Collection}
We instantiate agents on eight LLMs spanning closed- and open-weight families (Table~\ref{tab:base_models}).

\begin{wraptable}[14]{r}{0.45\textwidth}
   \centering
   \vspace{-15pt}
   \caption{Base LLMs used in our experiments.}
   \label{tab:base_models}
   \scriptsize
   \begin{tabular}{ll}
     \hline
     Model identifier        & Weight type   \\
     \hline
     Gemini-2.5-Flash-Lite   & Closed-weight \\
     Claude-3-Haiku          & Closed-weight \\
     GPT-4o-mini             & Closed-weight \\
     Qwen3-235B              & Open-weight   \\
     GPT-OSS-20B             & Open-weight   \\
     DeepSeek-3.1            & Open-weight   \\
     Llama-3.1-8B            & Open-weight   \\
     Devstral-Small-2505     & Open-weight   \\
     \hline
   \end{tabular}
   \vspace{-10pt}
 \end{wraptable}
Across all models and conditions, we collect 84{,}540 trajectories, with 99.98\% valid terminal actions. For each trajectory, we log step-level actions, step indices, (when applicable) observed peer actions, and internal tool use. For episodes that terminate early (claim the reward / "eat the marshmallow"), we additionally extract the final \texttt{Thought} trace immediately preceding the terminating action as the agent’s self-reported rationale.

From these logs we derive two primary data objects:
(1) time-to-event outcomes (survival status and time-to-claim), and
(2) termination rationales ($N{=}14{,}025$), one per early-terminated trajectory.

\subsection{Time-to-Event Outcomes}
We treat the first \textsc{Claim} as a discrete time-to-event outcome. For agent $i$, let $T_i$ denote the first step at which the agent claims (terminates) during the episode. If an agent never claims and reaches the horizon, the trajectory is right-censored at $T{=}20$. We analyze both terminal outcomes (claimed vs.\ censored) and the timing of claims across steps to distinguish early- versus late-stage failure risk.

\subsection{Survival and Hazard Modeling}
We report Kaplan-Meier survival curves \citep{kaplan1958nonparametric} for nonparametric comparisons across models and conditions. To quantify the association of experimental factors with time-varying termination risk, we fit a discrete-time hazard model implemented as a logistic regression on agent-step-level data. Let $h_i(t)$ be the hazard for agent $i$ at step $t$, i.e., the conditional probability of claiming at $t$ given survival up to $t-1$. The model is:
\begin{equation}
\text{logit}(h_i(t)) = \log\left(\frac{h_i(t)}{1-h_i(t)}\right) = \alpha_t + \mathbf{X}_i^T \boldsymbol{\beta},
\label{eq:hazard}
\end{equation}
where $\alpha_t$ is a set of step (time) dummies capturing baseline time effects, $\mathbf{X}_i$ encodes experimental condition indicators (and model family), and $\boldsymbol{\beta}$ are coefficients on the log-odds scale. We also report restricted mean survival time (RMST), defined as the area under the Kaplan-Meier curve up to horizon $T$, representing the average number of steps agents deferred before claiming. Full specification and implementation notes are given in Appendix~\ref{app:hazard_model}.

\subsection{Rationale Extraction (Pre-Claim Traces)}
For each trajectory that terminates early, we extract the final \texttt{Thought} trace produced immediately before the \textit{claim} (``eat'') action. These pre-claim traces are self-reported justifications aligned to an objective behavioral event (termination), which we analyze as behavioral text artifacts rather than as direct evidence of internal causal mechanisms.

\subsection{Taxonomy Induction and Labeling}
\paragraph{Taxonomy induction (human pilot, $N{=}100$).}
We developed a seven-category codebook via qualitative pilot analysis of 100 randomly sampled pre-claim traces. Two authors independently labeled these traces and reconciled disagreements by discussion, yielding the following categories:
\begin{enumerate}[leftmargin=*,noitemsep,topsep=0pt]
    \item \textbf{Cost-Benefit}: explicit trade-off rationale (e.g., expected value, probability, discounting)
    \item \textbf{Impulse/Craving}: immediate desire or temptation
    \item \textbf{Fatigue/Depletion}: exhaustion, boredom, or accumulated effort
    \item \textbf{Self-Control/Deontic}: permission, rules-as-self-regulation, or ``should/shouldn't'' framing
    \item \textbf{Rule Confusion}: misunderstanding or misbelief about task rules/state
    \item \textbf{Social Contagion}: peer behavior or norms as justification
    \item \textbf{Opportunity Framing}: reframing as a special chance or exception
\end{enumerate}

\paragraph{Large-scale classification ($N{=}14{,}025$).}
We classify all 14{,}025 pre-claim traces using Gemini 3.0 Flash with the fixed seven-category schema, instructing the model to select one category and to propose a novel label only when none applies.

\paragraph{Audit, agreement, and exclusions.}
Two authors audited a stratified sample of 200 traces (25 per category plus 25 from the novel/ambiguous pool), achieving Cohen's $\kappa=0.83$ against the LLM labels. Disagreements were resolved by discussion. Of the 14{,}025 traces, 13{,}780 (98.3\%) received one of the seven labels; the remaining 245 were flagged as novel or ambiguous and excluded from aggregate analyses. Class supports for the seven categories are reported in Appendix~\ref{app:annotation_statistics}.

\subsection{Linguistic and Consistency Features}
To characterize how expressed rationales vary across time, conditions, and model families, we extract surface-level features grounded in discourse analysis and modal semantics \citep{palmer2001mood,halliday2014introduction}. Table~\ref{tab:ling-features} summarizes the feature classes.

\begin{table}[t]
  \centering
  \small
  \begin{tabularx}{\columnwidth}{@{}llX@{}}
  \hline
  \textbf{Class} & \textbf{Feature} & \textbf{Lexical Items} \\
  \hline
  \multicolumn{2}{@{}l}{\textsc{Lexical}} & \\ 
  & Epistemic & might, could, perhaps, possibly, probably, maybe \\
  & Deontic & should, must, need, have to, ought \\
  & Immediacy & now, immediately, right now \\
  & Duration & already, waited, waiting, longer, still \\
  & Hedonic & want, desire, crave, temptation, enjoy \\
  & First-person & I, my, me \\
  \hline
  \multicolumn{2}{@{}l}{\textsc{Structure}} & \\ 
  & Causal & because, therefore, since, thus, hence \\
  & Conditional & if, would, could, unless \\
  & Contrastive & but, however, although, yet, despite \\
  \hline
  \multicolumn{2}{@{}l}{\textsc{Composite}} & \\
  & Argument density & causal + conditional + contrastive \\
  \hline
  \multicolumn{2}{@{}l}{\textsc{Consistency}} & \\
  & Self-contradiction & co-occurrence of wait-positive \emph{and} eat-positive statements \\
  \hline
  \end{tabularx}
  \caption{Linguistic features extracted from failure rationales. Lexical markers are normalized per 100 words. Argument density is the sum of the three reasoning structure rates.}
  \label{tab:ling-features}
  \end{table}

\textbf{Lexical markers} (per 100 words): epistemic modals (\emph{might, could, perhaps}), deontic modals (\emph{should, must, need}), immediacy terms (\emph{now, immediately}), duration terms (\emph{already, waited, still}), hedonic terms (\emph{want, desire, crave}), and first-person pronouns.

\textbf{Reasoning-marker density}: causal connectives (\emph{because, therefore}), conditional markers (\emph{if, would, unless}), and contrastive markers (\emph{but, however, although}). We define \emph{argument density} as the sum of these three rates.

\textbf{Intra-rationale inconsistency (minimal contradiction signal)}: co-occurrence of defer-positive (e.g., \emph{should wait, better to wait}) and claim-positive (e.g., \emph{claim now, I'll eat}) statements within a single rationale. This rule-based measure provides a conservative indicator of expressed inconsistency in deliberation text. We interpret it as correlational and diagnostic rather than causal.

\subsection{Appendix Pointers}
Appendix~\ref{app:pomdp} provides the full environment formalism. Appendix~\ref{app:hazard_model} provides additional hazard-model specification details. Appendix~\ref{app:marshmallow_background} provides background on delayed-gratification paradigms, and Appendix~\ref{app:annotation_statistics} reports annotation statistics.

\section{Results}

\subsection{Data Overview}

\begin{wraptable}{r}{0.7\textwidth}
  \centering
  \vspace{-15pt}
  \caption{Dataset overview (8 model families): counts, survival metrics, and data quality.}
  \label{tab:dataset-overview}
  \resizebox{\linewidth}{!}{
  \small
  \renewcommand{\arraystretch}{1.1}
  \begin{tabular}{@{}lrlrlr@{}}
  \hline  
  \multicolumn{2}{c}{\textbf{Data Summary}} & 
  \multicolumn{2}{c}{\textbf{Survival Metrics (overall)}} &
  \multicolumn{2}{c}{\textbf{Data Quality}} \\
  \hline
  Model Families               & 8                 & Initial Eat Rate        & 0.062  & Valid Rate            & 99.98\% \\
  Total Agent Trajectories     & 84{,}540          & Total Eat Rate          & 0.176  & Valid Trajectories    & 84{,}525 \\
  Total Cells                  & 512               & Winners Rate            & 0.824  & Invalid Trajectories  & 15 \\
  Risk Horizon ($T$)           & 20                & Median TTE (steps)      & 17.0   &                       &  \\
                               &                   & RMST (steps)            & 16.47  &                       &  \\
  \hline
  \end{tabular}
  }
  \vspace{-10pt}
  \end{wraptable}
 
We collected 84,540 agent trajectories across 8 model families and 64 experimental cells (Table~\ref{tab:dataset-overview}). Data quality was near-perfect (99.98\% valid). Of these, 14,025 agents chose to eat (17.6\%). We analyze their final \texttt{Thought} traces as failure rationales. After excluding 245 traces that did not fall under any of the seven taxonomy categories (novel or ambiguous), 13,780 labeled traces form the basis of our rationale analyses. 
\subsection{Failure Taxonomy: What Agents Say When They Fail}

\paragraph{Overall distribution.}
Failures are dominated by two categories: \textbf{Impulse/Craving} (37.4\%) and \textbf{Cost-Benefit} reasoning (34.4\%). Self-Control/Deontic Stance accounts for 12.8\%, Fatigue/Depletion for 10.6\%, with long-tail categories: Social Contagion (2.5\%), Rule Confusion (1.6\%), and Opportunity Framing (0.8\%).

\begin{figure}[t]
   \centering
   \begin{minipage}{0.48\textwidth}
    \includegraphics[width=\linewidth]{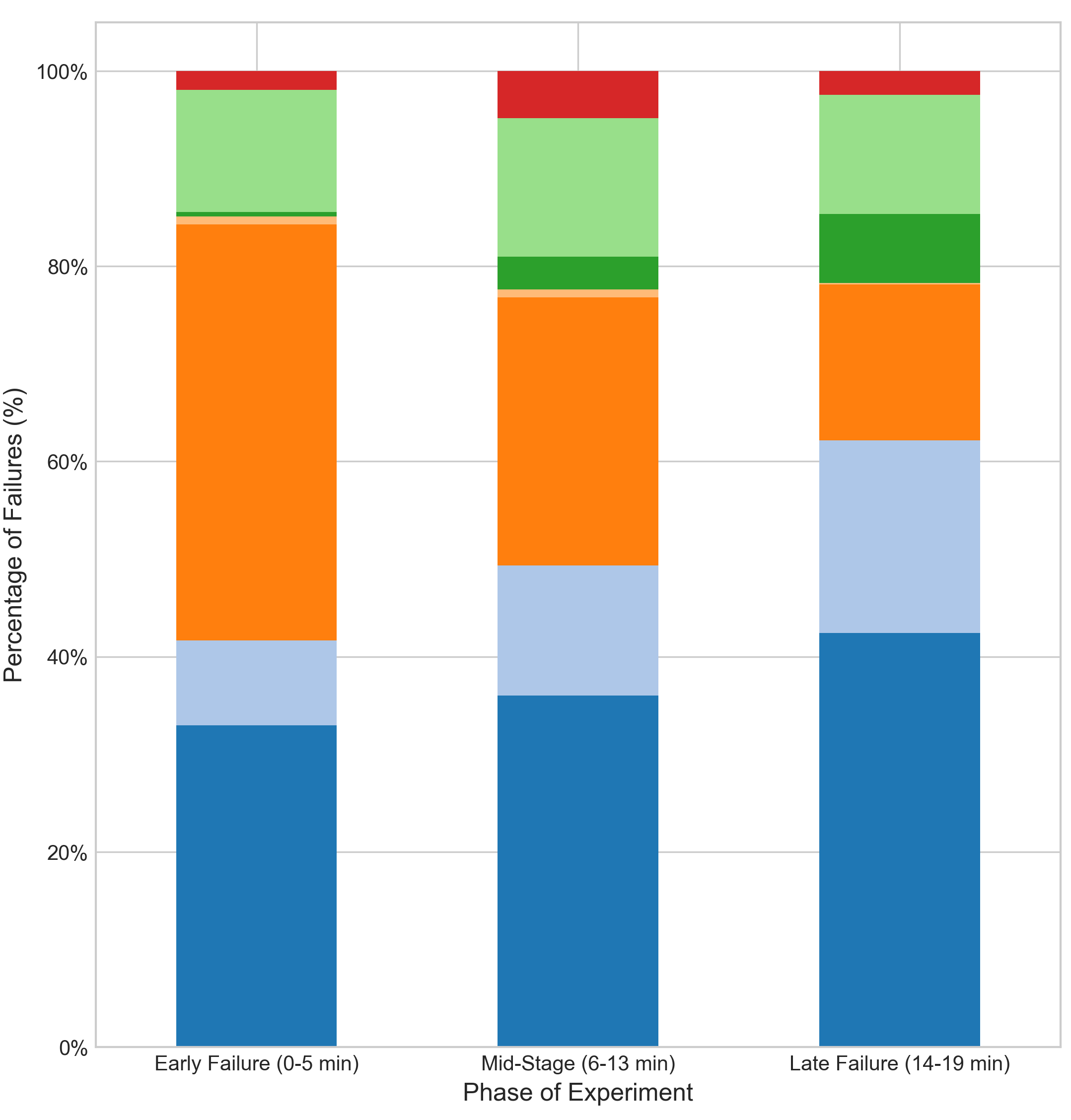}
    \caption{Temporal shift in failure rationales. Early failures (0-5 min) are dominated by Impulse/Craving (42.6\%), while late failures (14-19 min) show increased Cost-Benefit reasoning (42.4\%) and Fatigue/Depletion (19.7\%). Rule Confusion emerges primarily in late-stage failures.}
    \label{fig:temporal_rationales}
  \end{minipage}
  \hfill
  \begin{minipage}{0.48\textwidth}
    \centering
    \includegraphics[width=\linewidth]{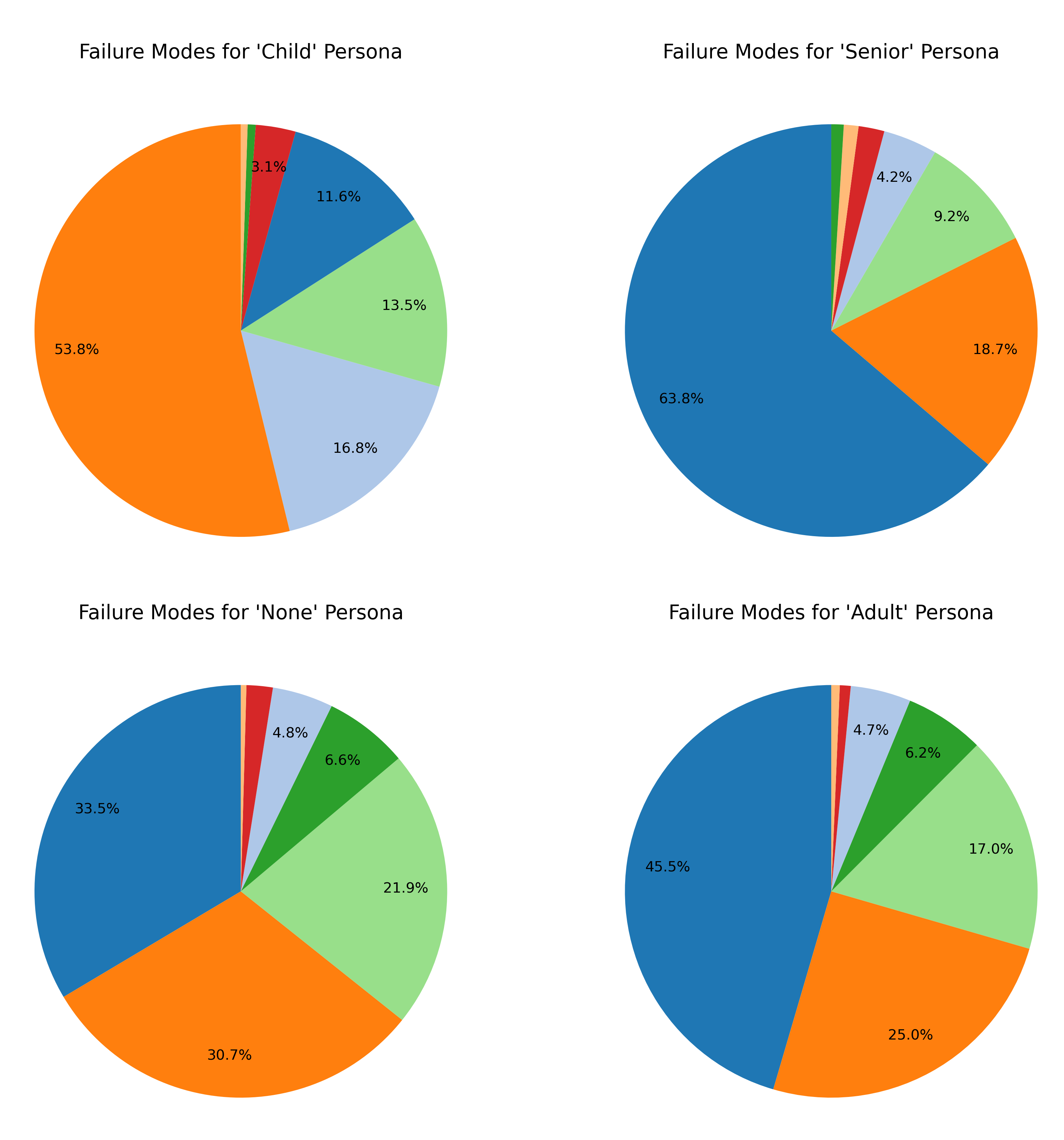}
    \caption{Dominant failure rationales by persona. Child personas are impulse-dominated (53.8\%), while senior personas show cost-benefit dominance (63.8\%). Adult and none personas exhibit intermediate, more balanced distributions.}
    \label{fig:persona_rationales}
  \end{minipage}
  \end{figure}

\paragraph{Temporal dynamics.}
The rationale mix shifts systematically across the experiment horizon (Figure~\ref{fig:temporal_rationales}). Early failures (first minutes, steps 0-5) are impulse-heavy (42.6\% Impulse/Craving). Late failures (last minutes, steps 14-19) show the inverse: Cost-Benefit dominates (42.4\%), Fatigue/Depletion surges to 19.7\%, and Impulse/Craving drops to 16.0\%.

\begin{figure*}[!t]
    \centering
    \includegraphics[width=0.9\textwidth]{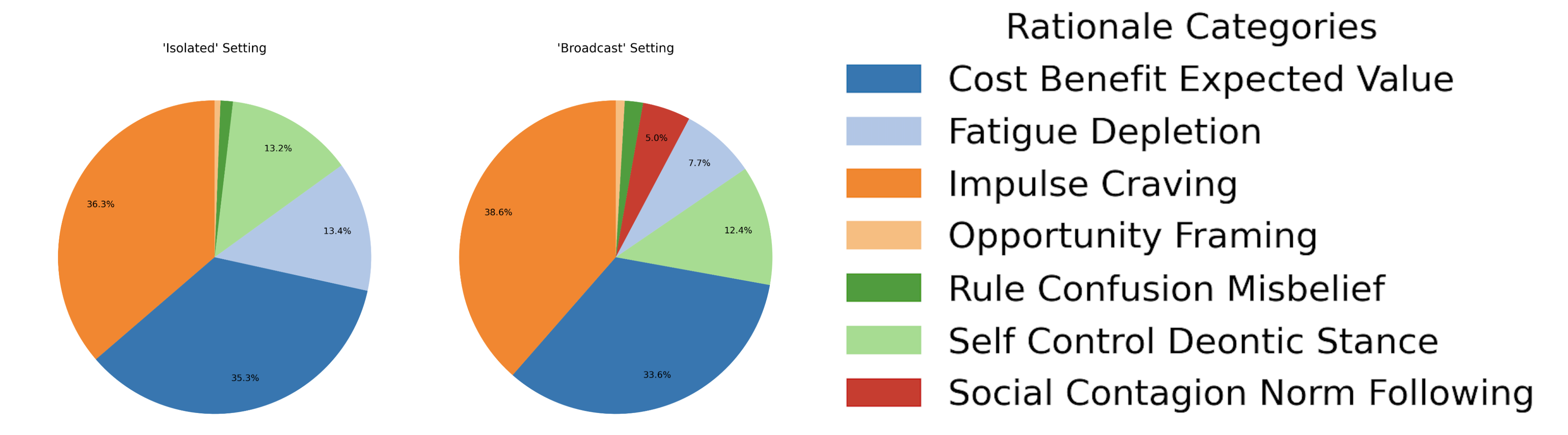}
    \caption{a. The impact of social setting on failure rationales. Broadcast condition elicits Social Contagion/Norm Following (5.0\%), while isolated conditions show elevated Fatigue/Depletion (13.4\% vs.\ 7.7\%). b. Rationale categories \& Color legend used for Figures 2-4}
    \label{fig:social_pies_legend}
    \end{figure*}

\paragraph{Social context shapes failure narratives.}
While broadcast versus isolated conditions yield near-zero effects on failure rates, they produce qualitatively different failure narratives (Figure~\ref{fig:social_pies_legend}). Broadcast uniquely elicits Social Contagion rationales (5.0\% vs.\ 0\% isolated), referring to agents citing peer behavior as justification. Isolated agents show elevated Fatigue/Depletion (13.4\% vs.\ 7.7\%), framing prolonged waiting as individual resource expenditure rather than norm deviation.

\paragraph{Persona-specific failure signatures.}
Age personas induce dramatically different rationale distributions (Figure~\ref{fig:persona_rationales}). Child personas exhibit a dominant impulse signature: 53.8\% Impulse/Craving, with Cost-Benefit at only 11.6\%. Senior personas show the inverse: 63.8\% Cost-Benefit, with Impulse/Craving at 18.7\%. These "failure fingerprints" indicate that models generate semantically coherent justifications matching assigned roles.

\paragraph{Tool policy effects.}
Mandatory deliberation (when agents MUST use the self-questioning tool) shifts rationales toward Cost-Benefit (36.1\% vs.\ 32.6\%) and away from Impulse/Craving (35.6\% vs.\ 39.4\%), suggesting forced self-questioning prompts explicit trade-off reasoning (Appendix~\ref{app:tool_policy}, Figure~\ref{fig:tool-policy-pies}).

\begin{figure*}[!t]
  \centering
  \includegraphics[width=0.9\textwidth]{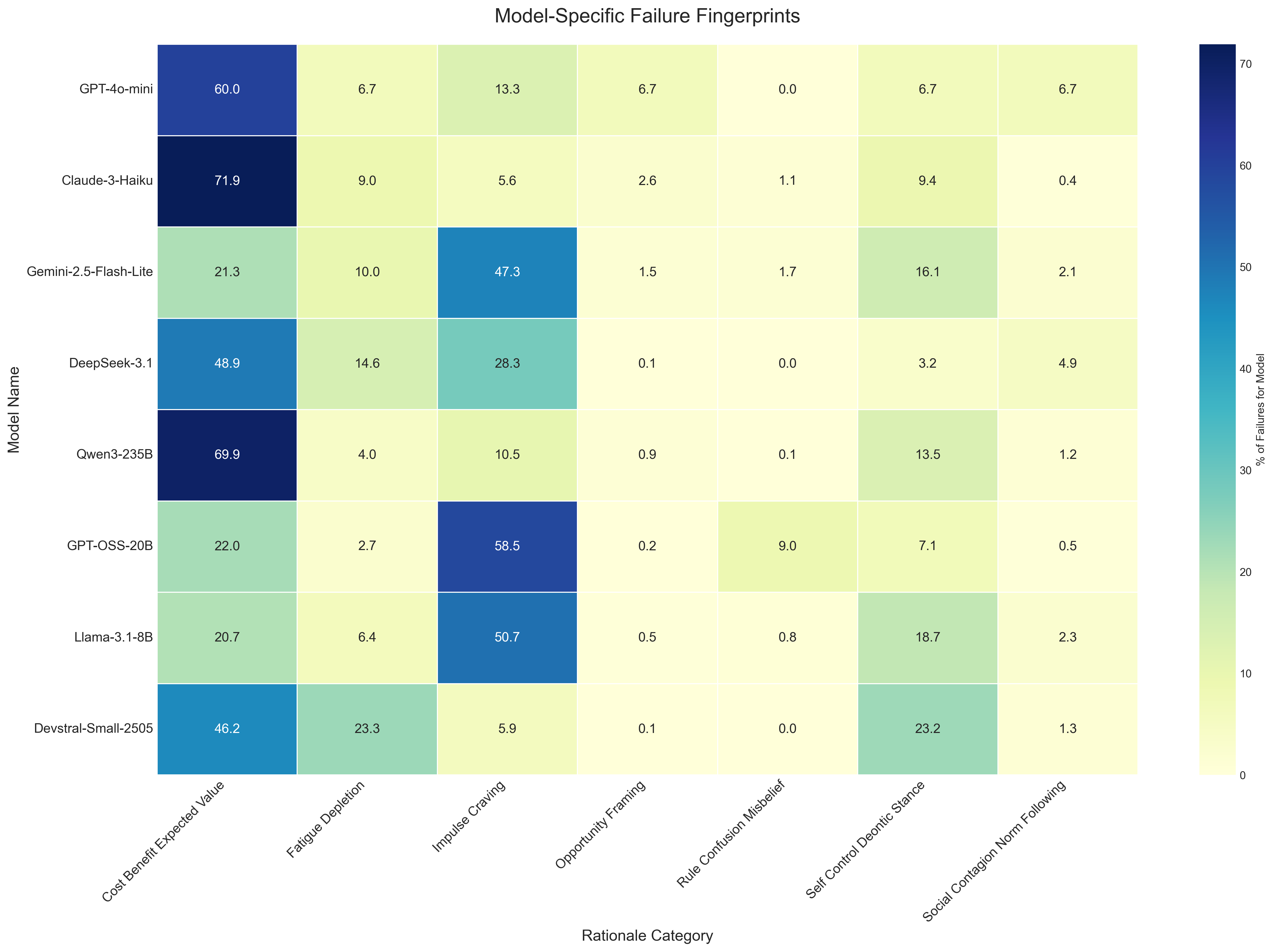}
  \caption{Model-specific failure fingerprints. Each row shows the distribution of rationale categories for a given model family. Cost-Benefit dominates in high-reliability models (Claude-3-Haiku, Qwen3-235B), while Impulse/Craving dominates in early-spike models (GPT-OSS-20B, Llama-3.1-8B).}
  \label{fig:model_fingerprints}
  \end{figure*}

\paragraph{Model-specific fingerprints.}
Different model families exhibit distinct rationale signatures (Figure~\ref{fig:model_fingerprints}). Cost-Benefit dominates in Claude-3-Haiku (71.9\%), Qwen3-235B (69.9\%), and GPT-4o-mini (60.0\%). Impulse/Craving dominates in GPT-OSS-20B (58.5\%), Llama-3.1-8B (50.7\%), and Gemini-2.5-Flash-Lite (47.3\%).

\subsection{The Deliberation-Inconsistency Association: More Reasoning, More Contradiction}

A natural assumption is that agents producing more elaborate reasoning should be more reliable. Our data suggests an opposite trend.

\paragraph{Reasoning density predicts contradiction, not success.}
\begin{wraptable}{r}{0.5\textwidth}
  \vspace{-20pt}
  \centering
  \small
  \caption{The deliberation-inconsistency association: self-contradiction rate increases with reasoning density.}
  \label{tab:deliberation_paradox}
  \begin{tabular}{lrr}
  \hline
  \textbf{Reasoning} & \textbf{Contradiction} & \textbf{$N$} \\
  \textbf{Density} & \textbf{\%} & \\
  \hline
  Q1 (lowest) & 0.43 & 3,479 \\
  Q2 & 1.54 & 3,452 \\
  Q3 & 1.15 & 3,404 \\
  Q4 (highest) & 1.89 & 3,445 \\
  \hline
  \multicolumn{3}{l}{\textit{Spearman $\rho$=0.040, $p<.001$}} \\
  \hline
  \end{tabular}
  \vspace{-10pt}
\end{wraptable}
Among failed agents, those producing more reasoning markers were more likely to exhibit self-contradictory rationales. We operationalize reasoning density as the sum of causal, conditional, and contrastive markers (per 100 words) and measure self-contradiction as co-occurrence of wait-positive and eat-positive statements within a single rationale.

Stratifying by reasoning density quartiles reveals a broadly increasing relationship (Table~\ref{tab:deliberation_paradox}): agents in the lowest quartile show 0.43\% contradiction rate, rising to 1.89\% in the highest quartile ($\rho{=}0.040$, $p{<}.001$). More reasoning is associated with more inconsistency, not less.

\paragraph{Temporal manifestation.}
Late-stage failures (minutes 14-19) exhibit significantly higher argument density than early failures (minutes 0-5): $M{=}3.74$ vs.\ $3.02$ ($t{=}-10.58$, $p{<}.001$). Yet self-contradiction (intra-rationale inconsistency) rates increase in parallel: 0.8\% (early) $\rightarrow$ 2.1\% (mid) $\rightarrow$ 2.8\% (late; $\chi^2{=}55.59$, $p{<}.001$). Agents who resist longer produce more elaborate reasoning and more internal inconsistency.

\paragraph{Implications.}
These findings speak directly to concerns about chain-of-thought faithfulness \citep{lanham2023cotfaithfulness,tutek2025measuring}. Densely elaborated traces should not be interpreted as evidence of reliable goal-directed behavior. For deployed agents, monitoring reasoning length or complexity is insufficient and potentially misleading, as a reliability diagnostic.

\subsection{Linguistic Dynamics of Failure}

\paragraph{Temporal shift in failure narratives.}
Failure rationales follow a temporal trajectory mirroring human self-regulation dynamics (Table~\ref{tab:hot_cool_shift}). \emph{Hot} features decrease over time: hedonic terms show $r{=}-0.254$ with minute ($p{<}.001$), declining from $M{=}4.98$ (early) to $M{=}2.48$ (late; $d{=}0.80$). \emph{Cool} features increase: duration terms show $r{=}0.409$ ($p{<}.001$), rising from $M{=}1.56$ to $M{=}3.97$ ($d{=}1.31$); deontic modals rise from $M{=}0.53$ to $M{=}1.48$ ($d{=}0.88$).

The ratio of immediacy-to-duration terms captures this shift: 1.70 in early failures (present-focused), declining to 0.63 in late failures (past-focused). First-person pronoun density increases significantly ($r{=}0.235$, $p{<}.001$), paralleling ego-depletion accounts in human self-control research \citep{baumeister2007strength}.

\begin{wraptable}{r}{0.55\textwidth}
  \vspace{-15pt}
  \centering
  \small
  \begin{tabular}{lrrrr}
  \hline
  \textbf{Feature} & \textbf{Early} & \textbf{Late} & \textbf{$r$} & \textbf{$d$} \\
  \hline
  \multicolumn{5}{l}{\textit{Hot features (decrease)}} \\
  \quad Hedonic & 4.98 & 2.48 & -0.254 & 0.80 \\
  \quad Immediacy & 2.67 & 2.52 & -0.019 & 0.06 \\
  \hline
  \multicolumn{5}{l}{\textit{Cool features (increase)}} \\
  \quad Duration & 1.56 & 3.97 & 0.409 & 1.31 \\
  \quad Deontic & 0.53 & 1.48 & 0.272 & 0.88 \\
  \quad First-person & 9.36 & 12.12 & 0.235 & 0.71 \\
  \hline
  \end{tabular}
  \caption{Hot-to-cool linguistic shift in failure narratives. Features are per 100 words, $r$ = Pearson correlation with minute, $d$ = Cohen's $d$ (early vs.\ late). All correlations significant at $p<.001$ except immediacy. The pattern mirrors human self-regulation dynamics: early failures are desire-laden, late failures are fatigue-framed.}
  \label{tab:hot_cool_shift}
  \vspace{-10pt}
  \end{wraptable}

\paragraph{Model-specific linguistic profiles.}
Linguistic profiles correspond to behavioral hazard regimes (Table~\ref{tab:ling-profiles}). High-reliability models (Claude-3-Haiku, GPT-4o-mini, Qwen3-235B) produce epistemically hedged rationales ($M{=}1.99$-$2.39$) with low contradiction rates (0.0-0.5\%). Early-spike models (GPT-OSS-20B, Gemini-2.5-Flash-Lite) show low epistemic hedging ($M{=}0.13$-$0.83$). Devstral-Small-2505, with bi-modal hazards, shows the highest contradiction rate (4.6\%) and elevated argument density ($M{=}4.09$).

\begin{table}[t]
  \centering
  \small
  \begin{tabular}{lrrrr}
  \hline
  \textbf{Model} & \textbf{Epist.} & \textbf{Arg.D.} & \textbf{Ctr.\%} & \textbf{$N$} \\
  \hline
  \multicolumn{5}{l}{\textit{High-reliability (near-flat hazard)}} \\
  \quad Claude-3-Haiku & 2.08 & 2.40 & 0.4 & 267 \\
  \quad GPT-4o-mini & 2.39 & 3.88 & 0.0 & 15 \\
  \quad Qwen3-235B & 1.99 & 3.41 & 0.5 & 934 \\
  \hline
  \multicolumn{5}{l}{\textit{Early-spike (impulsive failures)}} \\
  \quad Gemini-2.5-Flash-Lite & 0.83 & 2.77 & 0.5 & 4545 \\
  \quad GPT-OSS-20B & 0.13 & 2.36 & 1.1 & 1363 \\
  \hline
  \multicolumn{5}{l}{\textit{Bi-modal (late-stage vulnerable)}} \\
  \quad Devstral-Small-2505 & 1.23 & 4.09 & 4.6 & 1183 \\
  \quad Llama-3.1-8B & 0.59 & 4.37 & 1.2 & 2123 \\
  \hline
  \multicolumn{5}{l}{\textit{Context-sensitive}} \\
  \quad DeepSeek-3.1 & 0.99 & 2.98 & 1.5 & 3350 \\
  \hline
  \end{tabular}
  \caption{Model-specific linguistic profiles grouped by behavioral hazard regime. Epist.\ = epistemic modals; Arg.D.\ = argument density; Ctr.\ = self-contradiction rate. High-reliability models show epistemic hedging; late-stage-vulnerable models show elevated contradiction rates.}
  \label{tab:ling-profiles}
  \end{table}

  \subsection{Behavioral Grounding via Survival Profiles}

  The semantic categories are behaviorally grounded: they correspond to distinct temporal hazard profiles.
  
  \paragraph{Aggregate pattern.}
  The survival profile shows a characteristic shape: a sharp early impulse (6.2\% eat at minute 1) followed by a low-risk tail. Among the 17.6\% who fail, median time-to-eat is ${\approx}17$ minutes (RMST ${\approx}16.47$).
  
  \setlength{\columnsep}{10pt}
\begin{wraptable}{r}{0.6\textwidth}
  \centering
  \vspace{-15pt}
  \caption{Pooled hazard model (event-at-$t$). Odds ratios quantify factor effects on failure risk. Persona manipulations (age, such as \textit{child, senior}, or hedonic drive, e.g. \textit{crave}) strongly increase hazard. Mandatory deliberation (MUST) slightly increases risk.}
  \label{tab:hazard_pooled}
  \small
  \begin{tabular}{@{}lrrr@{}}
  \hline
  Contrast & $\beta$ & OR & $p$ \\
  \hline
  MUST vs MAY & 0.093 & 1.10 & $<.001$ \\
  Iso. vs Bcast & -0.009 & 0.99 & .514 \\
  H: like vs crave & -0.807 & 0.45 & $<.001$ \\
  H: neutral vs crave & -1.331 & 0.26 & $<.001$ \\
  H: none vs crave & -1.439 & 0.24 & $<.001$ \\
  A: child vs adult & 2.157 & 8.65 & $<.001$ \\
  A: senior vs adult & 1.723 & 5.60 & $<.001$ \\
  A: none vs adult & -0.102 & 0.90 & .022 \\
  \hline
  \end{tabular}
  \vspace{-10pt}
\end{wraptable}

\paragraph{Factor effects.}
A discrete-time hazard model confirms that persona manipulations strongly modulate failure risk (Table~\ref{tab:hazard_pooled}). Relative to adult, child increases hazard dramatically (OR${=}8.65$, $p{<}.001$), as does senior (OR${=}5.60$, $p{<}.001$). Mandatory tool use slightly increases hazard (OR${=}1.10$, $p{<}.001$), consistent with our finding that forced deliberation shifts rationales toward explicit trade-offs without improving outcomes.

\begin{figure}[h]
    \centering
    \includegraphics[width=0.4\textwidth]{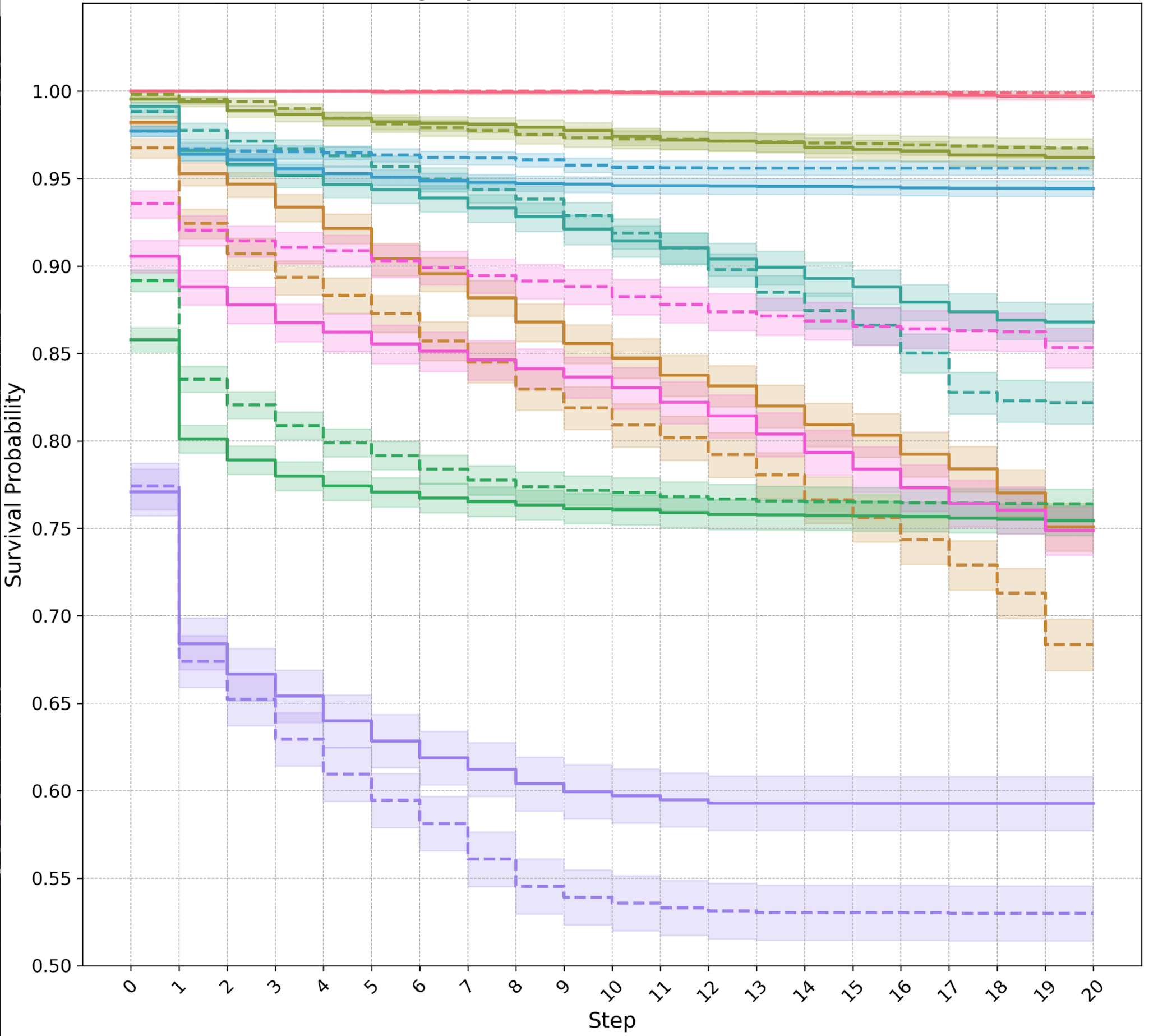}
    \includegraphics[width=0.4\textwidth]{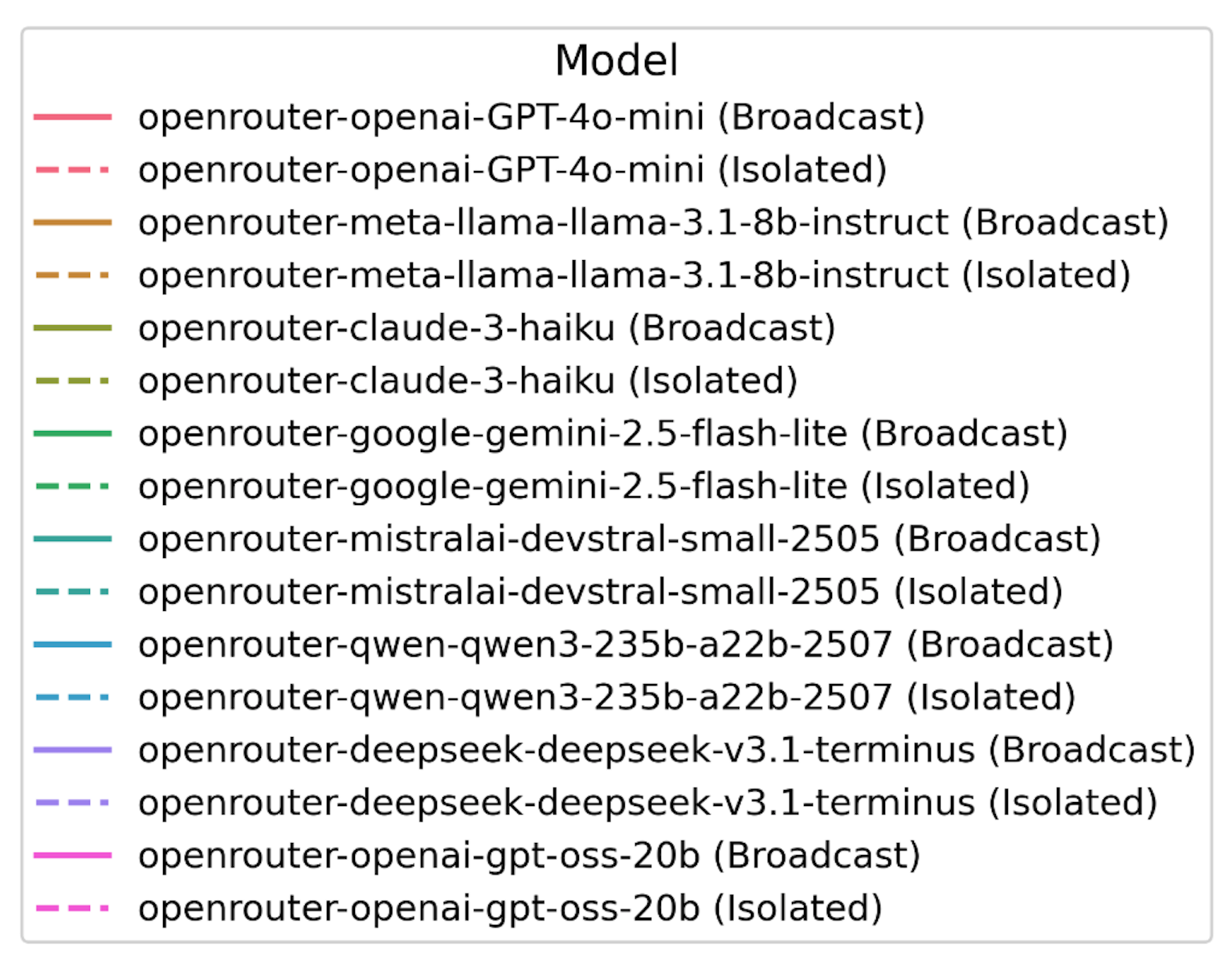}
    \caption{Kaplan-Meier survival curves for 8 models (broadcast vs.\ isolated), pooled over $N{=}84{,}540$ trajectories, illustrating three reliability regimes: Stable (Red), 
    Context Fatigue (Brown), and Impulse Failure (Purple).}
    \label{fig:km_all}
\end{figure}

  \paragraph{Three regimes.}
  Models cluster into three hazard regimes (Figure~\ref{fig:km_all}) that align with rationale fingerprints (Figure~\ref{fig:model_fingerprints}): (1) \emph{near-flat} profiles (GPT-4o-mini, Claude-3-Haiku) with Cost-Benefit-dominated rationales, (2) \emph{early-spike} profiles (Gemini, DeepSeek) with Impulse-dominated rationales, and (3) \emph{bi-modal} profiles (Llama-3.1-8B, Devstral-Small-2505) with elevated Fatigue rationales. This alignment validates that the semantic taxonomy captures behaviorally meaningful distinctions: models can be characterized not only by \emph{when} they fail but by \emph{how} they verbalize the decision to fail.

\section{Discussion}

Our findings have three implications for LLM agent deployment. First, more elaborated traces should not be treated as reliability signals. The deliberation-inconsistency association suggests that elaborate justifications accompany more self-contradiction, not less. Monitoring reasoning length or complexity may be misleading as a safety diagnostic. Second, model-specific failure fingerprints offer interpretable diagnostics beyond aggregate metrics: knowing that a model tends toward impulse-driven failures (early risk) versus fatigue-driven failures (late risk) can be potentially useful for deployment decisions and targeted interventions. Third, the systematic shift from impulse to fatigue rationales over time suggests that reliability monitoring should be time-aware, with different mitigation strategies for early versus late failures.

\paragraph{Limitations of the semantic analysis}
We treat rationale labels as diagnostics of expressed reasons, not claims about internal causal mechanisms. Chain-of-thought traces may not faithfully reflect underlying computations \citep{lanham2023cotfaithfulness,tutek2025measuring}. However, anchoring rationales to objective behavioral events (claiming ("eating") vs.\ deferring ("waiting")) constrains interpretation: regardless of internal causality, these labels characterize the decision narratives agents produce when abandoning tasks.
The hot-to-cool shift most likely arises because LLMs have learned the linguistic patterns humans use when describing self-regulation, not because the models undergo genuine resource depletion. When placed in a structurally similar situation, they reproduce those patterns. The task itself also constrains what justifications are plausible: early in the horizon, few temporal cues are available, so desire language dominates. Later, accumulated duration and effort cues make fatigue framing natural. Which of these two factors impacts the shift, learned convention or task structure, remains an open question.

\paragraph{Limitations of the survival analysis} 
Success in this controlled setting is necessary but not sufficient for real-world reliability: we do not evaluate adaptive replanning, tool use in open-ended environments, or domain expertise. \textbf{Personas as stressors:} Persona prompts are controlled stylizations that can surface prioritization tradeoffs and consistency failures under role constraints. Differences across persona conditions should not be interpreted as stable properties of demographic groups and they may reflect prompt-induced objectives rather than intrinsic model traits. \textbf{Constraints:} Our fixed decoding (temperature=0.5) and binary action space prioritize experimental control over ecological breadth.
\noindent\textbf{Scope of the setting.} This delayed-reward design isolates a minimal commitment pressure under systematic perturbations, but it does not model open-world uncertainty or complex action spaces. Its value is as a controlled consistency stress test enabling time-to-event modeling and failure-rationale cartography, not as a full deployment proxy.

 \paragraph{Future work.}
We consider three extensions. First, testing whether the observed failure fingerprints and inconsistency signals generalize to richer agentic tasks with larger action spaces and explicit multi-question consistency requirements. Second, investigating whether rationale-aware prompting or self-verification can reduce contradiction rates and shift time-to-event profiles toward more stable regimes. Third, developing lightweight linguistic monitors that flag elevated inconsistency risk (e.g., contradiction cues, low epistemic hedging) in deployed multi-turn systems.

\section{Ethical considerations}

All authors have read and will adhere to the ICLR Code of Ethics.\footnote{\url{https://iclr.cc/public/CodeOfEthics}} This study evaluates synthetic interactions among large language models in controlled environments. It involves no human participants, no personally identifiable data, and no collection of user data; as such, it did not require institutional ethics review at our institution. Persona prompts and social-exposure conditions are used solely as experimental stylizations to induce controlled variation in agent behavior; we do not target or stereotype real demographic groups, and results should not be interpreted as claims about humans. All third-party models and APIs were used in accordance with their terms and licenses. We are unaware of conflicts of interest that could bias this work and will disclose any that arise.

\bibliography{iclr2026_conference}
\bibliographystyle{iclr2026_conference}

\appendix
\section{Appendix}

\section{Environment Formalism}\label{app:pomdp}
We formalize the interaction as a finite-horizon Partially Observable Markov Decision Process (POMDP) \citep{Kaelbling1998} with horizon $T{=}20$. At each step $t\in\{1,\dots,T\}$, the environment is in state $S_t$ and the agent receives an observation
\begin{equation}
O_t = [\text{Time}(t),\; \mathds{1}_{\text{bc}}\cdot \texttt{peer}_t], \label{eq:obs}
\end{equation}
where $\texttt{peer}_t$ summarizes recent peer actions in broadcast conditions and is empty in isolated conditions ($\mathds{1}_{\text{bc}}=0$). The agent may optionally invoke an internal deliberation tool \texttt{raise\_a\_question} up to a per-step cap $C$; these tool calls do not alter the environment state.

After optional tool use, the agent emits a constrained action
\begin{equation}
A_t \in \{\textsc{Defer},\textsc{Claim}\},
\end{equation}
implemented as the strings $\{\text{"I wait"},\text{"I eat the marshmallow"}\}$, respectively. Choosing \textsc{Claim} terminates the episode immediately with an immediate payoff (+1). Choosing \textsc{Defer} advances the episode to the next step. Agents that reach the horizon without \textsc{Claim} receive the delayed payoff (+2). Formally, the interaction loop at each step is:
\begin{align}
O_t &= [\text{Time}(t),\mathds{1}_{\text{bc}}\cdot\texttt{peer}_t], \notag\\
A_t &= \pi_{\theta}\!\left(O_t,\{\texttt{Q}(O_t,i)\}_{i=1}^{k_t}\right), \quad \text{s.t. } 0\le k_t\le C, \notag\\
R_{t+1}, S_{t+1}, O_{t+1} &= \mathcal{E}(S_t, A_t), \notag
\end{align}
where $\texttt{Q}$ denotes \texttt{raise\_a\_question}. In isolated conditions, observations are fully determined by time, reducing the process to a finite-horizon MDP \citep{Puterman1994}.

\section{Hazard Model Specification}\label{app:hazard_model}
We estimate a discrete-time hazard model as logistic regression on agent-step-level data. Let $T_i$ denote the first step at which agent $i$ selects \textsc{Claim}. For each trajectory, we construct a row for each step $t$ up to the event or censoring. Define the event indicator
\[
y_{i,t} =
\begin{cases}
1, & \text{if } t=T_i \text{ (first \textsc{Claim})},\\
0, & \text{if } t<T_i \text{ (still \textsc{Defer})},
\end{cases}
\]
and for trajectories that never claim, we set $y_{i,t}=0$ for all $t\in\{1,\dots,T\}$ and treat them as right-censored at $T$.

Let $h_i(t)=\Pr(T_i=t \mid T_i \ge t, \mathbf{X}_i)$ be the discrete-time hazard at step $t$. The model is:
\begin{equation}
\text{logit}(h_i(t)) = \log\left(\frac{h_i(t)}{1-h_i(t)}\right) = \alpha_t + \mathbf{X}_i^T \boldsymbol{\beta}, \label{eq:hazard_appendix}
\end{equation}
where $\alpha_t$ is a set of step (time) dummies capturing baseline time effects, $\mathbf{X}_i$ is a vector of covariates encoding experimental condition indicators (e.g., broadcast vs.\ isolated, persona attributes, deliberation policy) and model family, and $\boldsymbol{\beta}$ are coefficients on the log-odds scale. Coefficients are interpreted as associations with the time-varying probability of \textsc{Claim} conditional on survival up to step $t$.

In addition to regression estimates, we report Kaplan-Meier survival curves \citep{kaplan1958nonparametric}. We also report the restricted mean survival time (RMST), defined as the area under the Kaplan-Meier survival curve up to horizon $T$, representing the average number of steps agents defer before claiming.

  \subsection{Dataset scale, Base models and Headline Statistics}\label{app:scale}
Table~\ref{tab:overall_metrics} summarizes global statistics computed from the included CSVs.
  
  \begin{table}[h]
  \centering
  \caption{Overall summary (all 8 model families).}
  \label{tab:overall_metrics}
  {\tiny
  \begin{tabular}{l r}
  \hline
  Initial eat rate & 0.0616 \\
  Winners rate & 0.8245 \\
  Median time-to-eat (steps) & 16.9785 \\
  RMST (steps) & 16.4716 \\
  Valid outcomes (N) & 84525 \\
  Invalid outcomes (N) & 15 \\
  Data quality rate & 0.9999 \\
  \hline
  \end{tabular}
  }
  \end{table}
  
  We instantiate our ReAct agents on eight base LLMs, spanning both open- and closed-weight APIs. All experiments share the same environment, prompts, and decoding setup across models (see Methods). The models are summarized in Table~\ref{tab:base_models_appendix}.
  
  \begin{table}[h!]
    \centering
    \caption{Base LLMs used in our experiments.}
    \label{tab:base_models_appendix}
    \begin{tabular}{ll}
      \hline
      Model identifier        & Weight type   \\
      \hline
      Gemini-2.5-Flash-Lite   & Closed-weight \\
      Claude-3-Haiku          & Closed-weight \\
      GPT-4o-mini             & Closed-weight \\
      Qwen3-235B              & Open-weight   \\
      GPT-OSS-20B             & Open-weight   \\
      DeepSeek-3.1            & Open-weight   \\
      Llama-3.1-8B            & Open-weight   \\
      Devstral-Small-2505     & Open-weight   \\
      \hline
    \end{tabular}
  \end{table}

  \noindent Communication outcomes by social condition (mean across cells) are in Table~\ref{tab:comm_means}.
  
  \begin{table}[h]
  \centering
  \caption{Broadcast vs.\ Isolated (means across cells).}
  \label{tab:comm_means}
  \begin{tabular}{l r r r}
  \hline
  Condition & Winners rate & Median TTE & RMST \\
  \hline
  Broadcast & 0.8272 & 16.97 & 16.48 \\
  Isolated  & 0.8218 & 16.99 & 16.47 \\
  \hline
  \end{tabular}
  \end{table}

  \section{Prompts and Experimental Materials (Verbatim excerpts)}\label{app:prompts}
  \subsection*{Environment YAML template (excerpt)}
  {\small
  \begin{verbatim}
max_steps: 20
question_budget: 3

rules_prompt:
  - You can choose to eat your single 
    marshmallow at any time by saying
    "I eat the marshmallow".
  - If you eat the marshmallow your 
    waiting is over.
  - However, if you can wait for 
    {max_steps} minutes without eating 
    it, you will receive 2 marshmallows.

answer_format: >
  CRITICAL RULE: In your JSON 
  {Answer: <value>} you must only 
  produce one of: "I eat the 
  marshmallow" | "I wait".
  \end{verbatim}
  }
  
  \subsection*{Reasoning/tool policy prompt (excerpt)}
  {\small
  \begin{verbatim}
<loop_rules>
CRITICAL TOOL USE INSTRUCTION:
YOU {tool_use_policy} USE TOOLS 
TO VALIDATE YOUR DECISIONS
{visible_question_cap}
</loop_rules>

<tool_list>
raise_a_question ... You have a 
limited number of questions you 
can ask.
</tool_list>
  \end{verbatim}
  }
  
  \subsection*{Design factors and decoding}
  \noindent Factors (full matrices and seeds in the ZIP):
  {\small
  \begin{itemize}
    \item \textbf{Social:} communication\_type $\in$ \{broadcast, isolated\}
    \item \textbf{Policy:} tool\_use\_policy $\in$ \{may, must\}
    \item \textbf{Persona (hedonic):} \{crave, like, neutral, none\}
    \item \textbf{Persona (age):} \{child, adult, senior, none\}
    \item \textbf{Visibility toggles:} visible\_question\_budget=false, distraction=none
    \item \textbf{Decoding:} temperature=0.5, top\_p=0.7, max\_tokens=1000
    \item \textbf{Global caps:} max\_steps=20, question\_cap\_per\_minute=3, reasoning\_step\_cap\_total=10
  \end{itemize}
  }
  
  \section{Agent Loop and Validation Rules}\label{app:loop}
  \paragraph{Validation rules.}
  Only two canonical actions are valid at each minute: \texttt{"I eat the marshmallow"} or \texttt{"I wait"}.
  Non-canonical terminal strings that imply success (e.g., \texttt{"I won"}) are normalized to \texttt{waited\_full} at $t{=}19$ (steps are indexed 0-19) unless an eat-event was recorded earlier.
  
  \paragraph{Loop pseudocode.}
  {\small
  \begin{verbatim}
for t in {0..19}:
  prompt = build_prompt(system, 
    persona, policy, minute=t, 
    social_context, history)
  reply = model(prompt, 
    temperature, top_p, max_tokens)
  action = normalize(reply)
    # "I eat" | "I wait" | misc.
  if action == "I eat the 
    marshmallow": 
      record event at t; break
  elif action == "I wait": 
      record censor at t; continue
  If no eat at t=19, 
    mark waited_full
  \end{verbatim}
  }
  
  \subsection{Example Trajectories (verbatim excerpts)}\label{app:traj}
  \noindent We show short, verbatim excerpts for orientation.
  
  \vspace{2mm}
  \noindent\textbf{Example A} (\texttt{GPT-4o-mini}, MAY, adult/neutral, isolated):
  
  {\small
  \begin{verbatim}
t=0 user: "No time has passed 
  yet... You are in a room alone"
t=0 assistant: Thought: ... 
  Tool: raise_a_question: What 
  would be the appropriate 
  strategy?  PAUSE
t=1 user: "Minute 1 of 20 has 
  passed... What will you do?"
t=1 assistant: Thought: ... 
  Tool: raise_a_question: ...
...
t=17 assistant: Thought: ... 
  {{Answer: "I eat the 
  marshmallow"}}
  \end{verbatim}
  }
  
  \vspace{2mm}
  \noindent\textbf{Example B} (\texttt{Qwen3-235B}, MUST, child/crave, broadcast):
  
  {\small
  \begin{verbatim}
t=0 user: "No time has passed 
  yet... You are in a room 
  together."
t=0 assistant: Thought: ... 
  Tool: raise_a_question: ...
t=1..19 assistant: ... 
  {{Answer: "I wait"}}
Final: normalized as 
  waited_full at t=19
  \end{verbatim}
  }

  \section{Hazard Shape Regimes}\label{app:hazard_regimes}

  We identify three distinct hazard-shape regimes across model families, each corresponding to different failure dynamics and rationale signatures (Section 4.5). These regimes are visualized by plotting instantaneous hazard rates (probability of eating at time $t$, conditional on surviving to $t$) across experimental conditions.
  
\begin{figure*}[t]
    \centering
    \begin{subfigure}[b]{0.6\textwidth}
        \centering
        \includegraphics[width=\textwidth]{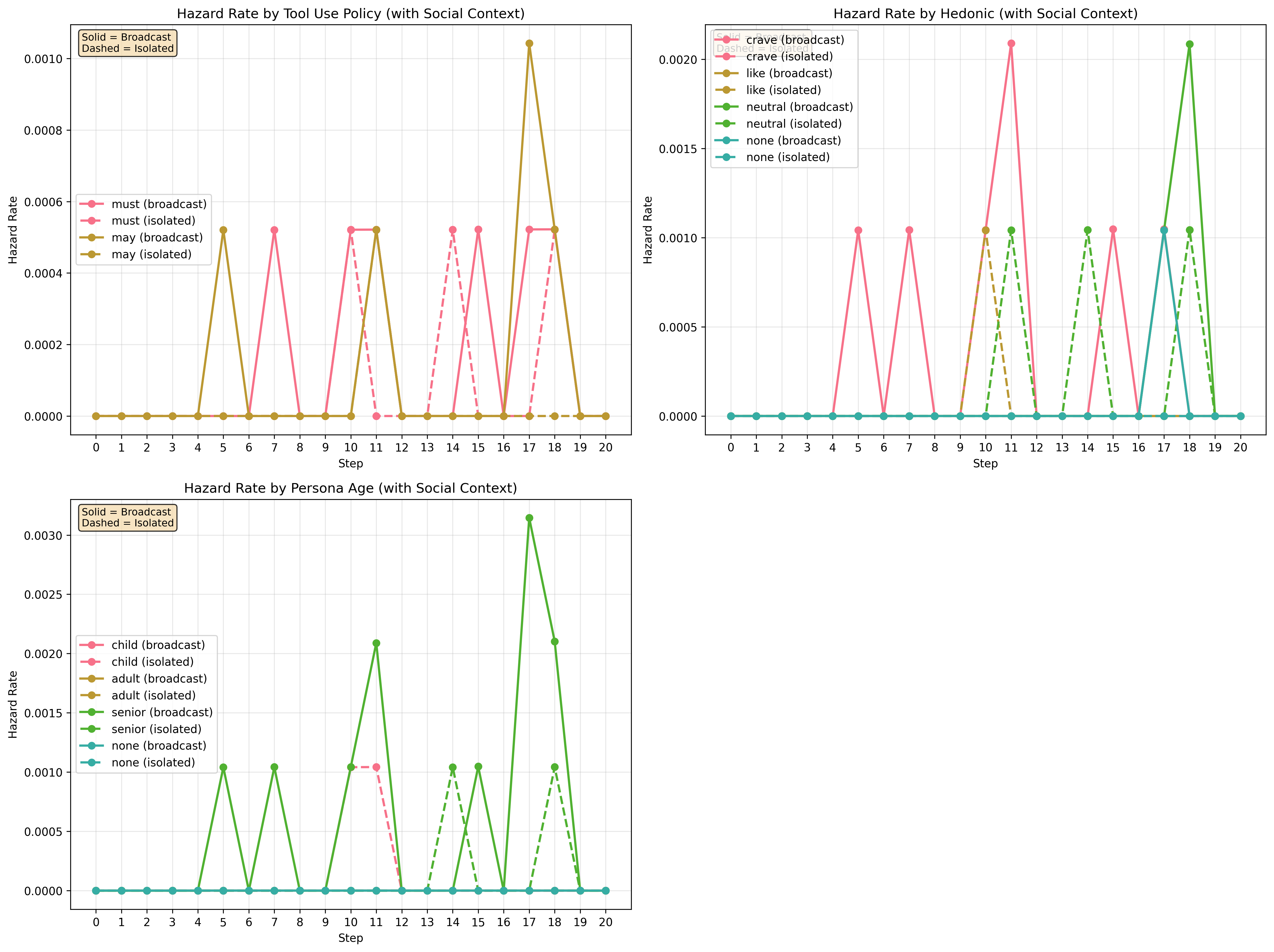}
        \caption{near-flat}
        \label{fig:spike_flat}
    \end{subfigure}
    \par\vspace{0.5cm}
    \begin{subfigure}[b]{0.6\textwidth}
        \centering
        \includegraphics[width=\textwidth]{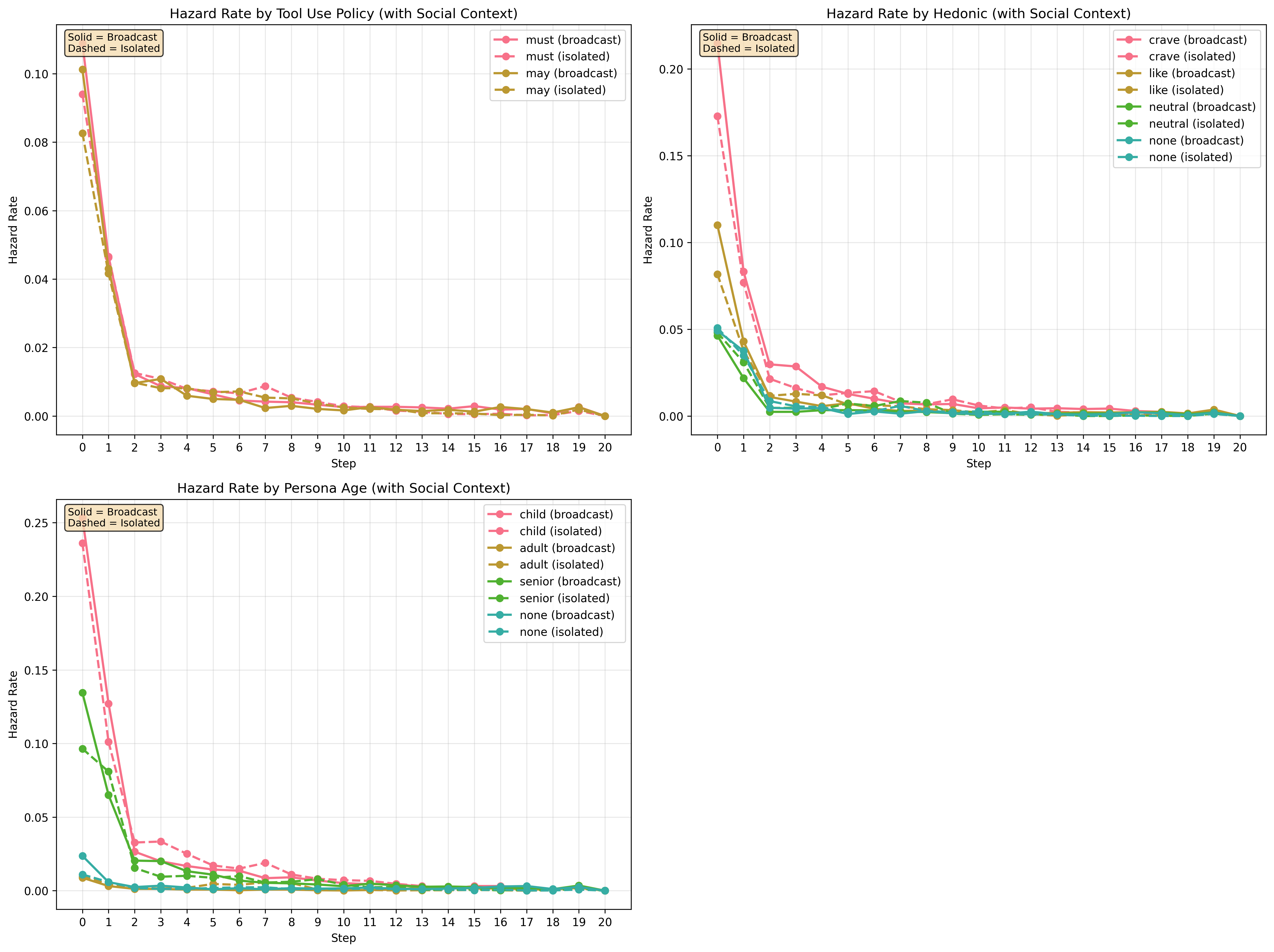}
        \caption{early spike}
        \label{fig:spike_left}
    \end{subfigure}
    \par\vspace{0.5cm}
    \begin{subfigure}[b]{0.6\textwidth}
        \centering
        \includegraphics[width=\textwidth]{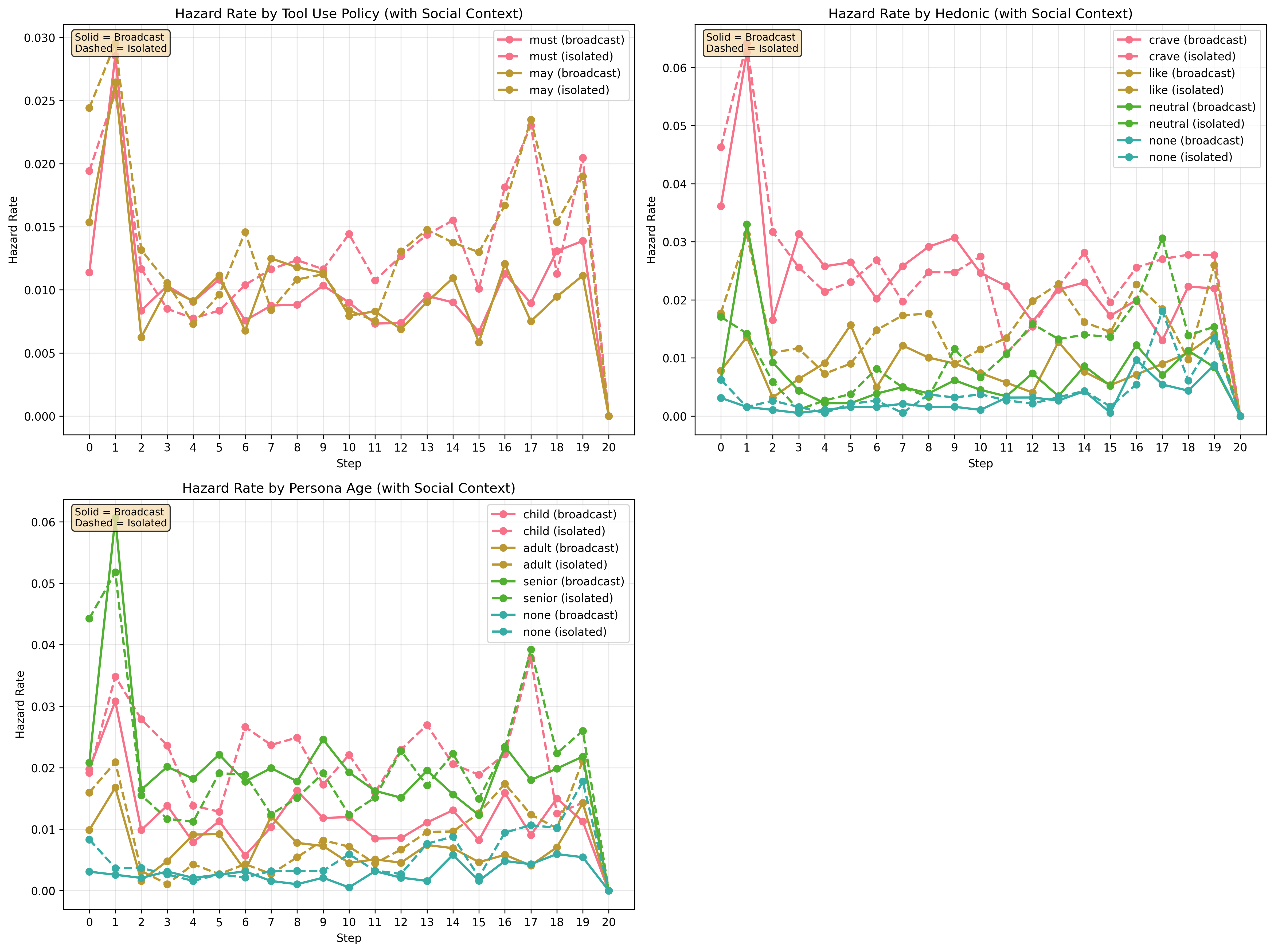}
        \caption{early and late spike}
        \label{fig:two_spikes}
    \end{subfigure}
    \caption{Three distinct hazard-shape regimes across model families. Panel (a) shows a \textbf{near-flat} profile (e.g., GPT-4o-mini) with consistently low risk. Panel (b) shows an \textbf{early spike} (e.g., Gemini, Qwen) where failure risk is concentrated in the first few minutes. Panel (c) shows a \textbf{bi-modal} profile (e.g., Llama-3.1) exhibiting both an initial impulse and a late-stage rise in hazard.}
    \label{fig:spikes}
  \end{figure*}

  \textbf{Near-flat} (Figure~\ref{fig:spike_flat}): Models like GPT-4o-mini exhibit consistently low hazard throughout the horizon, with occasional late-stage spikes under specific persona conditions. These models produce predominantly Cost-Benefit rationales when they do fail.
  
  \textbf{Early-spike} (Figure~\ref{fig:spike_left}): Models like Gemini-2.5-Flash-Lite and Qwen3-235B show hazard concentrated in the first 2-3 minutes, with child and crave personas amplifying the initial spike. Failures are Impulse/Craving-dominated.
  
  \textbf{Bi-modal} (Figure~\ref{fig:two_spikes}): Models like Llama-3.1-8B and Devstral-Small-2505 exhibit both an early impulse and a secondary late-stage rise (minutes 14-19). These models show elevated Fatigue/Depletion rationales and the highest self-contradiction rates.
  
\subsection{Tool Policy Effects on Rationales}\label{app:tool_policy}

\begin{figure*}[t]
  \centering
  \includegraphics[width=0.9\textwidth]{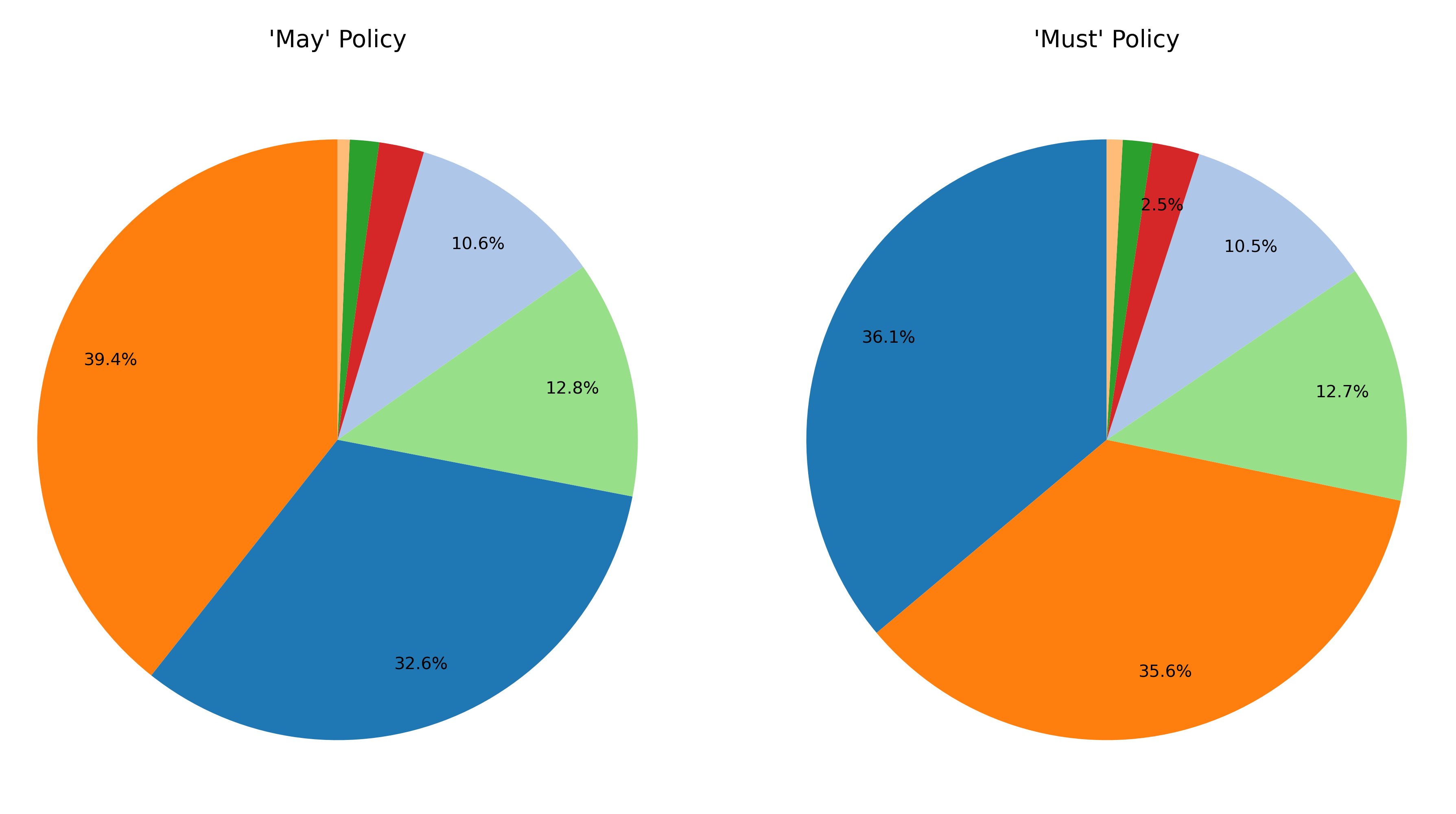}
  \caption{How tool policy influences failure rationales. Mandatory deliberation (MUST) shifts rationales toward Cost-Benefit reasoning (36.1\% vs.\ 32.6\%) and away from Impulse/Craving (35.6\% vs.\ 39.4\%). Despite more explicit trade-off reasoning, mandatory deliberation slightly increases failure risk.}
  \label{fig:tool-policy-pies}
\end{figure*}

As noted in Section 4.2, mandatory deliberation (MUST policy) shifts the distribution of failure rationales compared to optional tool use (MAY policy). Figure~\ref{fig:tool-policy-pies} visualizes this effect.

Under the MAY policy, Impulse/Craving dominates (39.4\%), with Cost-Benefit reasoning at 32.6\%. When deliberation is mandatory (MUST), this pattern partially inverts: Cost-Benefit rises to 36.1\% while Impulse/Craving drops to 35.6\%. Self-Control/Deontic Stance and Fatigue/Depletion remain relatively stable across conditions (12.8\% vs.\ 12.7\% and 10.6\% vs.\ 10.5\%, respectively).

This shift suggests that forced self-questioning prompts agents to articulate explicit trade-off reasoning rather than acting on immediate desire. However, as reported in the main text, mandatory deliberation paradoxically increases overall failure risk (OR=1.10, $p<.001$). The deliberation requirement appears to focus attention on the temptation rather than away from it, which is consistent with the "hot/cool" model where salient cue attention increases impulsive responding, even when that attention is framed as deliberation.

\section{Annotation Statistics}\label{app:annotation_statistics}

\begin{table}[h!]
  \centering
  \small
  \begin{tabular}{lrr}
  \hline
  \textbf{Rationale category} & \textbf{Count} & \textbf{\% of labeled} \\
  \hline
  Impulse\_Craving & 5{,}158 & 37.4 \\
  Cost\_Benefit\_Expected\_Value & 4{,}744 & 34.4 \\
  Self\_Control\_Deontic\_Stance & 1{,}759 & 12.8 \\
  Fatigue\_Depletion & 1{,}455 & 10.6 \\
  Social\_Contagion\_Norm\_Following & 342 & 2.5 \\
  Rule\_Confusion\_Misbelief & 218 & 1.6 \\
  Opportunity\_Framing & 104 & 0.8 \\
  \hline
  \end{tabular}
  \caption{Support per rationale category for labeled traces ($N=13,780$). Novel/ambiguous traces ($N=245$) are excluded from these counts.}
  \label{tab:label-support}
  \end{table}

\section{Ablations}\label{app:ablations}
  
\begin{figure*}[t]
\centering
\includegraphics[width=\textwidth]{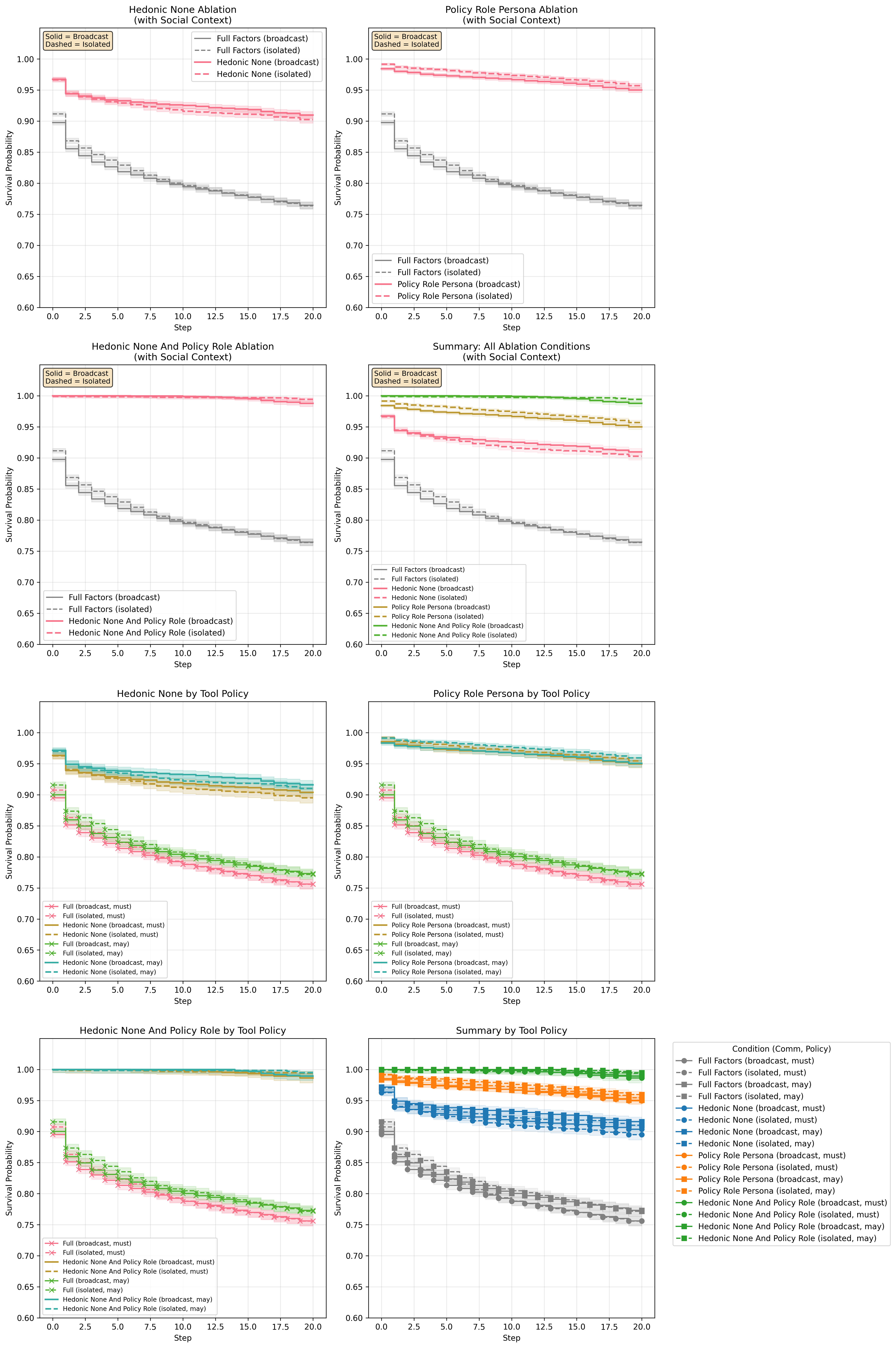}
\caption{Data pooled across all 8 model families. Kaplan-Meier survival by ablation condition. Persona removals suppress the early spike and flatten the hazard across the horizon.}
\label{fig:ablation_km}
\end{figure*}

\begin{figure*}[t]
\centering
\includegraphics[width=0.8\textwidth]{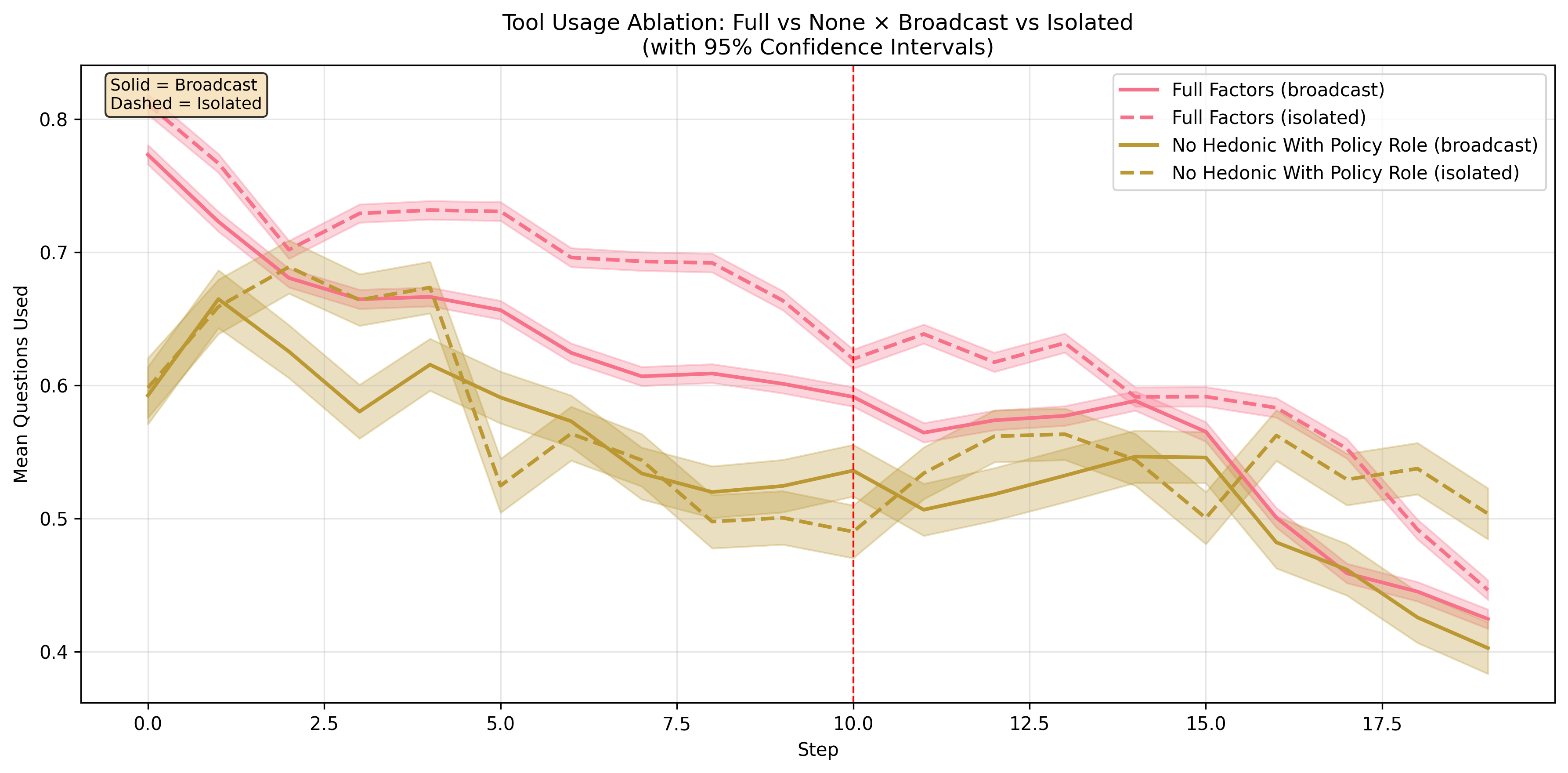}
\caption{Data pooled across all 8 model families. Tool-use under ablations: mean questions per step (with 95\% CIs) for Full vs.\ None across social conditions. Lower question rates accompany improved survival under persona removals.}
\label{fig:ablation_tool_usage}
\end{figure*}

\begin{figure*}[t]
\centering
\includegraphics[width=0.8\textwidth]{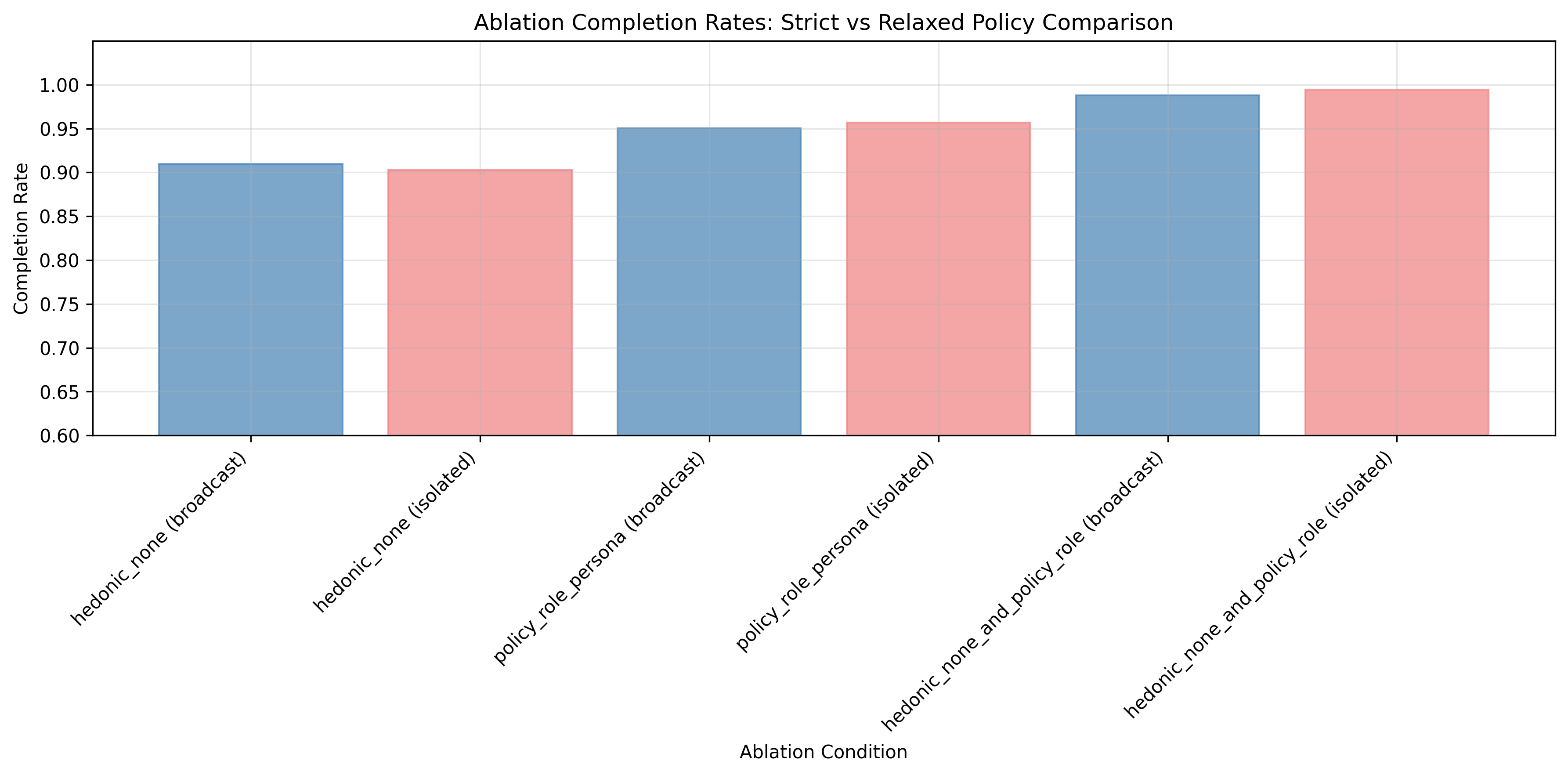}
\caption{Data pooled across all 8 model families. Alternative completion comparison (strict vs.\ relaxed policy view) across ablations. Results mirror the main figure: the combined removal delivers the highest completion in both social conditions.}
\label{fig:ablation_completion_alt}
\end{figure*}

\paragraph{Additional diagnostics.}
Figure~\ref{fig:ablation_km} shows Kaplan-Meier survival under each ablation, confirming that persona removals suppress the early spike and yield flatter hazards throughout the horizon. Figure~\ref{fig:ablation_completion_alt} provides a compact completion comparison (strict vs.\ relaxed policy view) consistent with the main text. Figure~\ref{fig:ablation_tool_usage} reports question-tool dynamics: ablations lower per-step question rates, while the \textsc{must} policy maintains higher usage and corresponds to higher hazard, matching our pooled hazard estimates.

\begin{figure*}[t]
\centering
\includegraphics[width=0.8\textwidth]{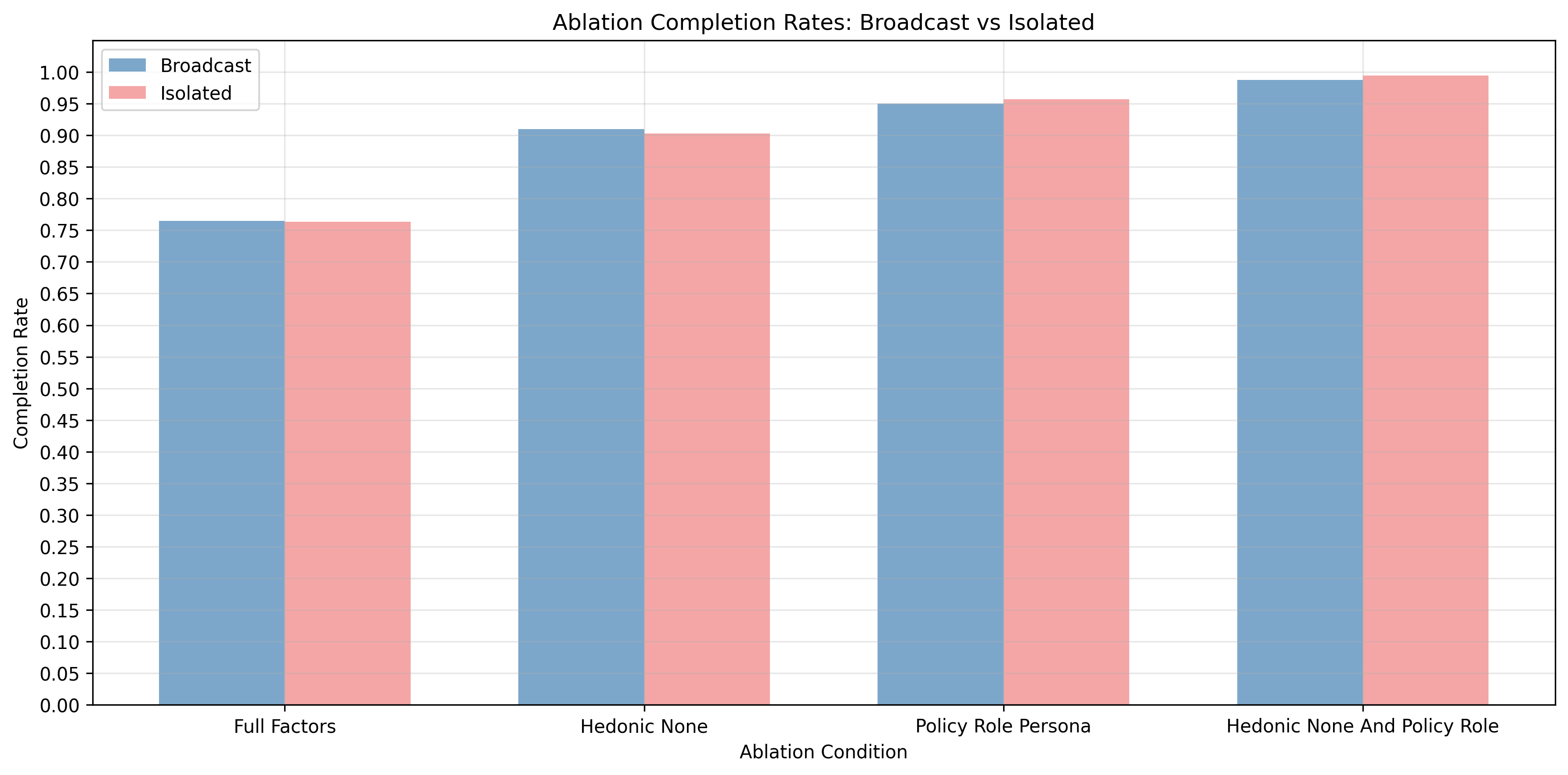}
\caption{Data pooled across all 8 model families. Completion by ablation condition and social visibility. Removing personas (hedonic, policy-role) increases completion. The combined removal approaches 1.0 and compresses the broadcast-isolated gap.}
\label{fig:ablation_completion_main}
\end{figure*}

  \section{Background on the Stanford marshmallow experiment}
  \label{app:marshmallow_background}
  
  The classic delay-of-gratification paradigm was introduced in a series of studies at Stanford, often
  referred to as the "marshmallow test" \citep{mischel1972marshmallow}. In the canonical setup, preschool children were seated alone in a room with a single, visible treat (e.g., a marshmallow) and
  told that they could either ring a bell or call the experimenter back at any time and consume that
  treat immediately, or wait for a fixed delay to receive a larger reward (typically two treats). The primary behavioral measure was the amount of time children waited before choosing the immediate reward, or whether they successfully waited until the experimenter returned.
  
  Follow-up experiments systematically varied the attentional and cognitive context of the task. For example, children were instructed to think about the treat in concrete "hot" terms (e.g., its taste and smell) or in abstract, "cool" terms referring to its shape or an imagined picture, or were given distractions to shift attention away from the reward. These manipulations showed that cool terms or redirecting attention increase waiting time, whereas focusing on the immediate reward decreases
  it, motivating the hot/cool model of self-control \citep{metcalfe1999hot}.
  
  Later work examined the stability and interpretation of individual differences in waiting. Longitudinal studies initially suggested that longer waiting times predicted a range of later-life outcomes, but subsequent work showed that these links are substantially moderated by environmental reliability and socioeconomic context \citep{kidd2013rational, watts2018revisiting}. Neural studies in adults further implicated prefrontal circuitry and control networks in intertemporal choice and self-control \citep{casey2011behavioral}.
  
  Our benchmark abstracts away many complexities of the human setting (e.g., no uncertainty about reward delivery, no rich social or familial context) while preserving the core structure: at each discrete time step, an agent must choose between an immediate smaller reward and a delayed larger reward. We adapt this structure to a discrete-time survival-analysis frame for LLM agents, with
  explicit manipulation of social exposure, internal state prompts (personas), and self-questioning policy.

  \section{Reproducibility}  \label{app:reproducibility}

  We provide comprehensive methodological details to support independent reimplementation. Prompt templates are reproduced verbatim in Appendix~\ref{app:prompts}, and the paper reports the full factorial design (conditions, levels, and sampling), outcome definitions, and labeling guidelines used for the failure-rationale taxonomy. We will release trajectory logs and annotations (including failure-rationale labels and extracted linguistic features) as structured files, along with aggregated summaries used in the paper.

  \section{Supplementary Figures}\label{app:figures}

This section collects additional figures referenced in the main text or generated by the analysis scripts. All figures are reproducible from the scripts in the \texttt{reproduce\_analysis.zip} file.




\begin{figure}[h]
  \centering
  \includegraphics[width=.95\linewidth]{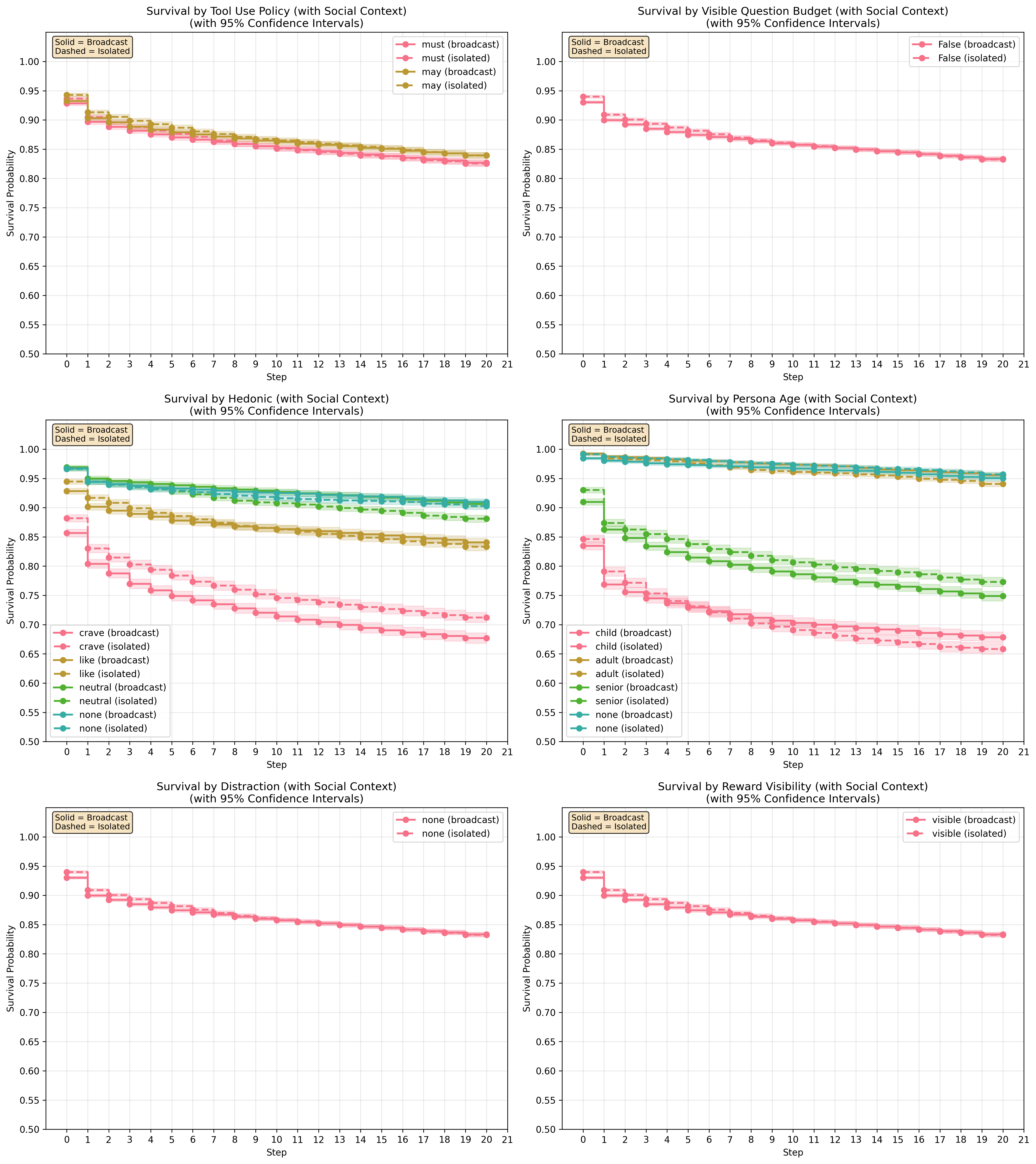}
  \caption{Kaplan-Meier survival across all families.}
  \label{fig:app_km_all}
  \end{figure}
  
  \begin{figure}[h]
  \centering
  \includegraphics[width=.95\linewidth]{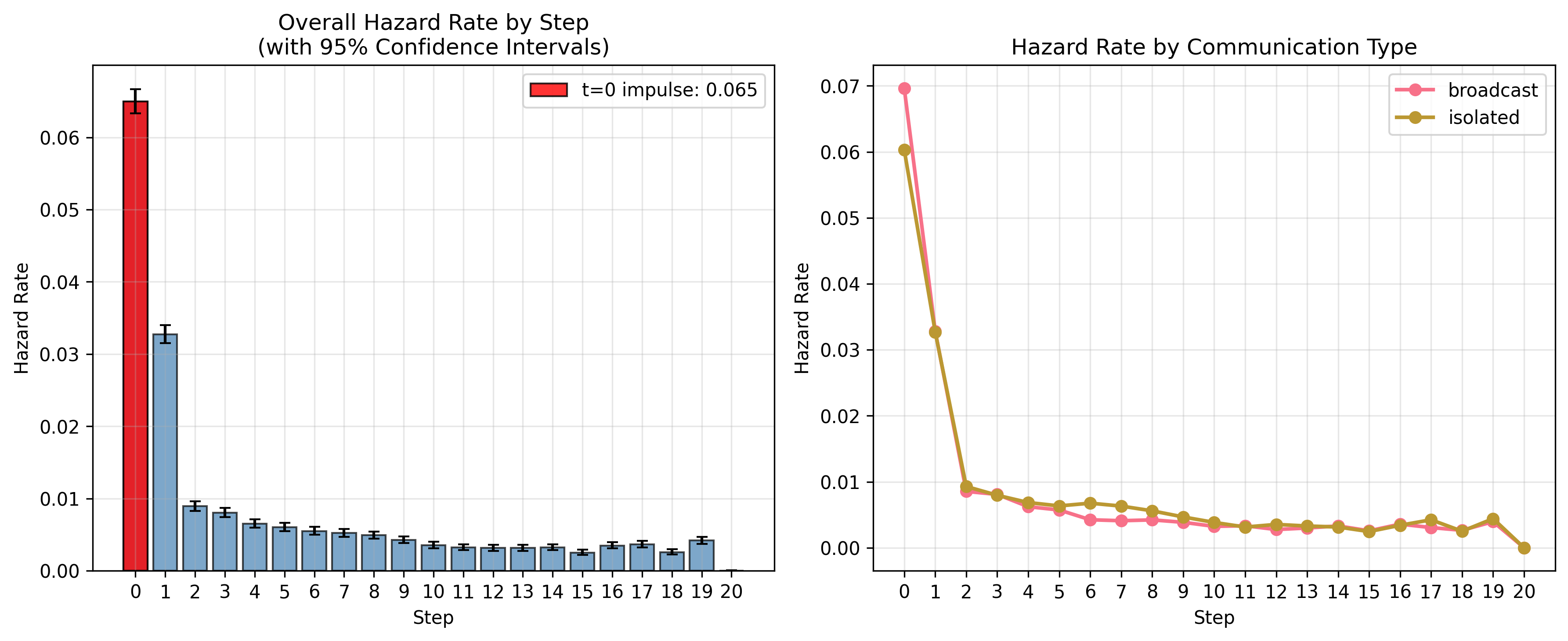}
  \caption{Discrete-time hazard by minute (pooled).}
  \label{fig:app_hazard_rates}
  \end{figure}
  
  \begin{figure}[t]
  \centering
  \includegraphics[width=0.9\columnwidth]{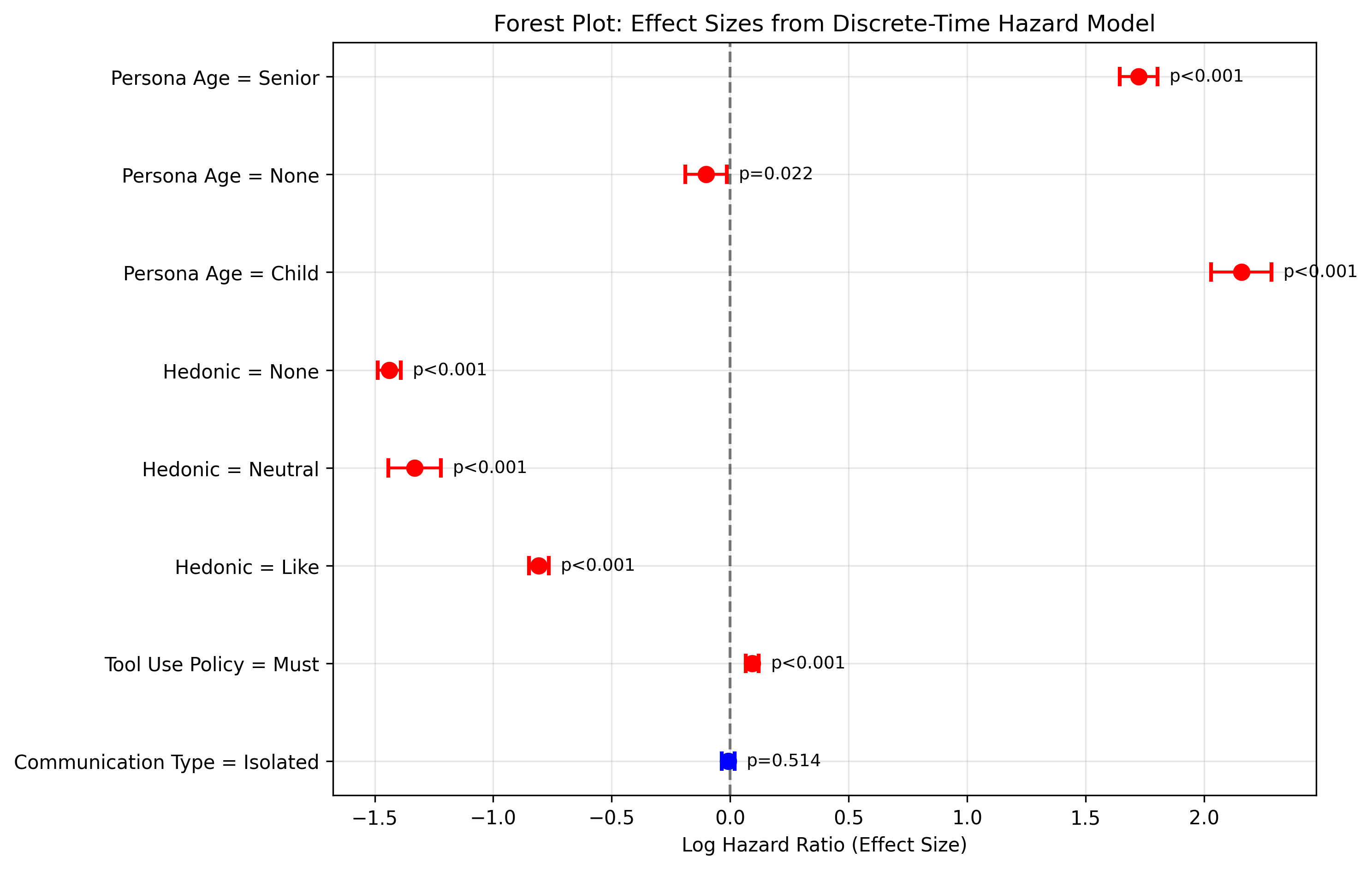}
  \caption{Data pooled across all 8 model families. Effect-size forest plot (pooled ORs with 95\% CIs) showing the impact of experimental factors on the hazard of eating. Note the strong increase in risk for \textit{child} and \textit{senior} personas, and the risk reduction for \textit{neutral} drive.}
  \label{fig:forest_plot}
  \end{figure}

  \begin{figure}[t]
    \centering
    \includegraphics[width=0.4\textwidth]{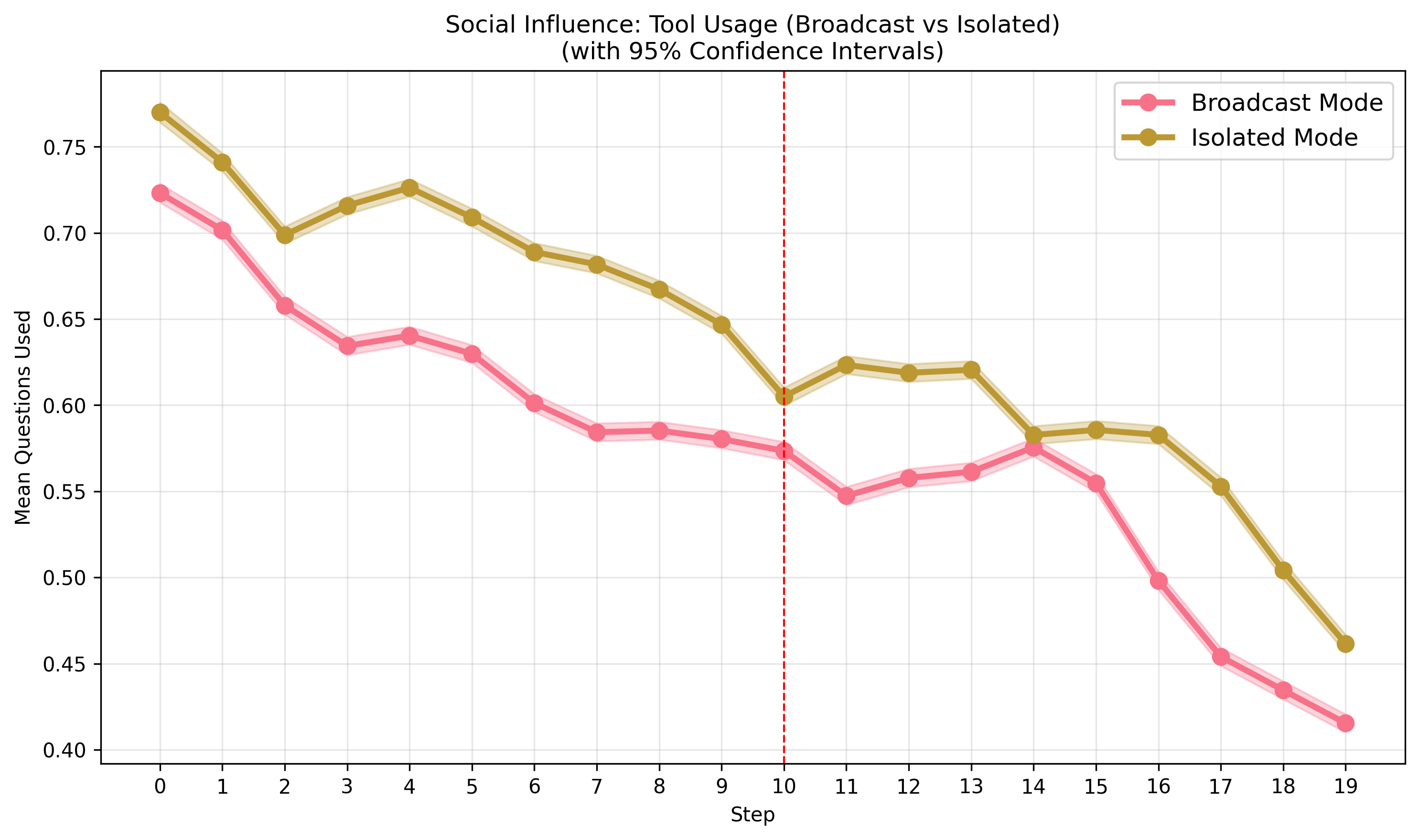}
    \caption{Data pooled across all 8 model families. Mean questions per step with 95\% CIs, split by social visibility (broadcast vs.\ isolated). Rates decline over time in both conditions and are close when pooled, consistent with the near-zero broadcast main effect on hazard.}
    \label{fig:tool_usage_patterns}
    \end{figure}


    \begin{figure*}[t]
      \centering
      \begin{subfigure}[b]{0.32\textwidth}
          \centering
          \includegraphics[width=\textwidth]{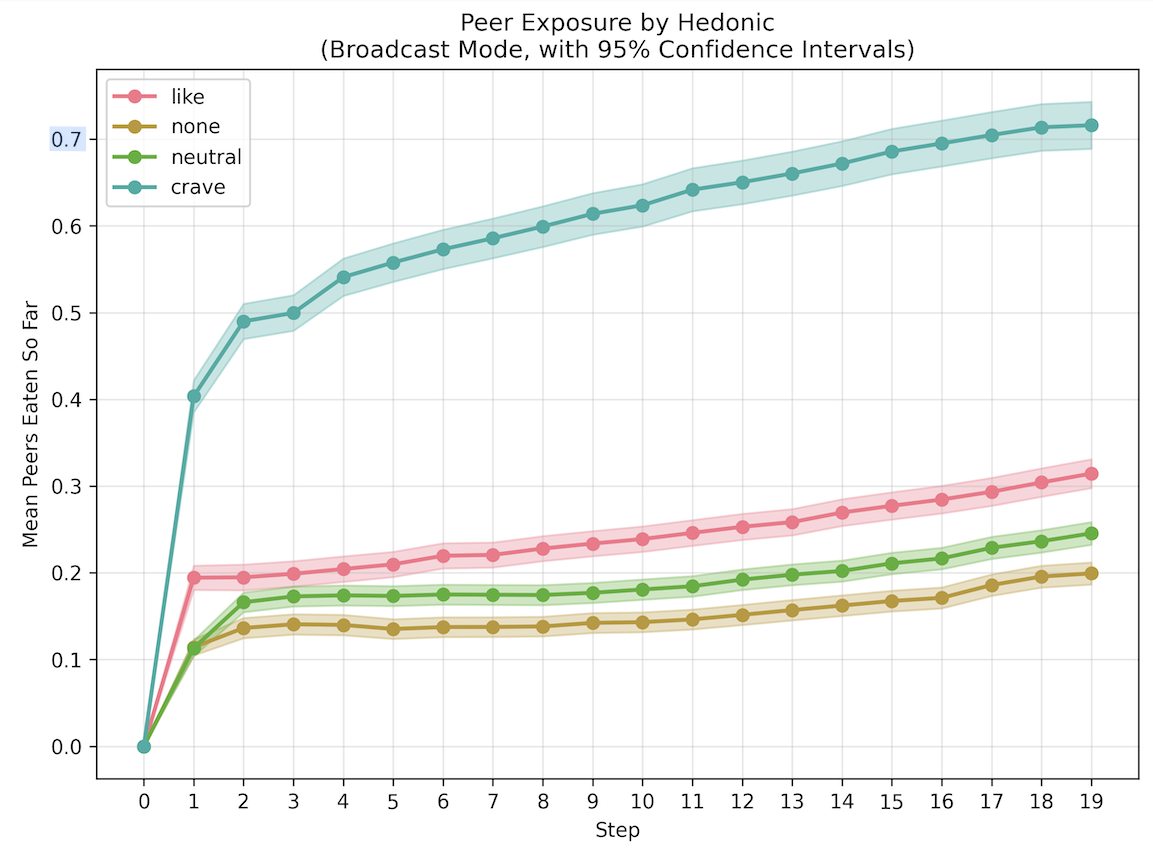}
          \caption{PE by hedonic}
          \label{fig:peer_exp_by_drive}
      \end{subfigure}
      \hfill
      \begin{subfigure}[b]{0.32\textwidth}
          \centering
          \includegraphics[width=\textwidth]{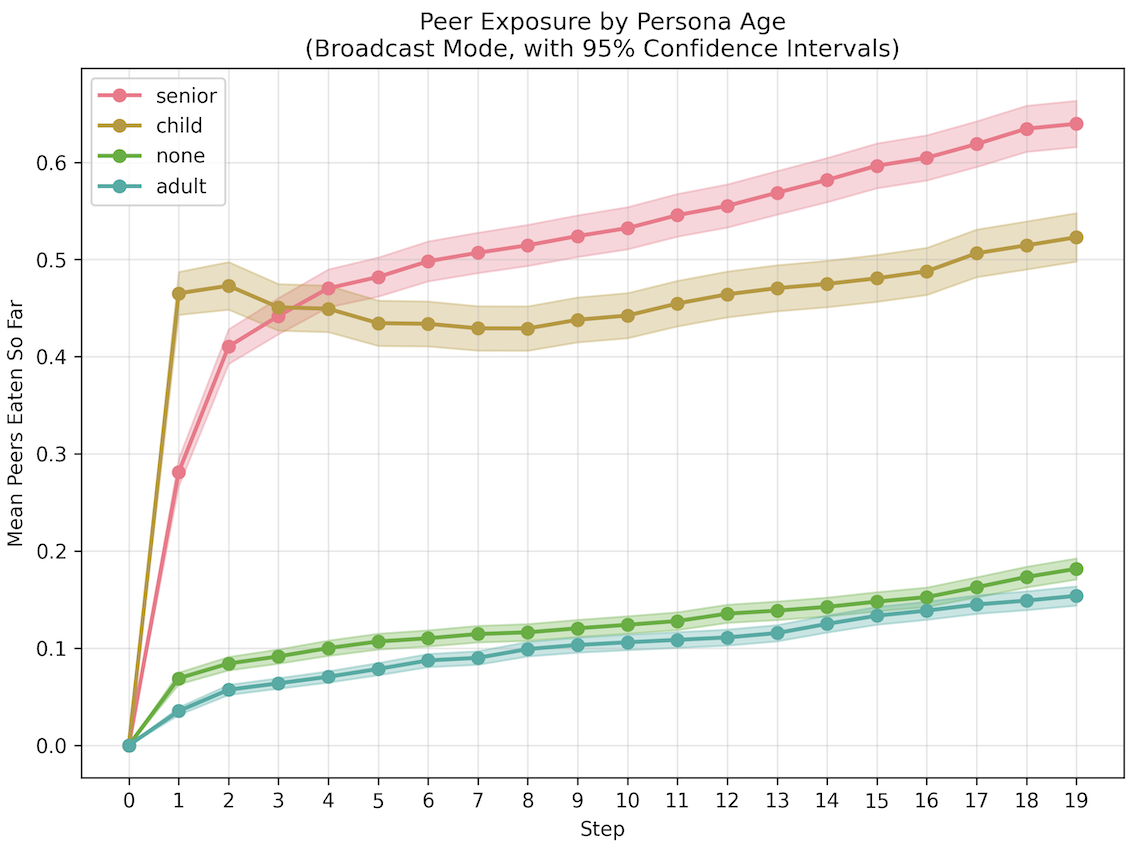}
          \caption{PE by age}
          \label{fig:peer_exp_by_age}
      \end{subfigure}
      \hfill
      \begin{subfigure}[b]{0.32\textwidth}
          \centering
          \includegraphics[width=\textwidth]{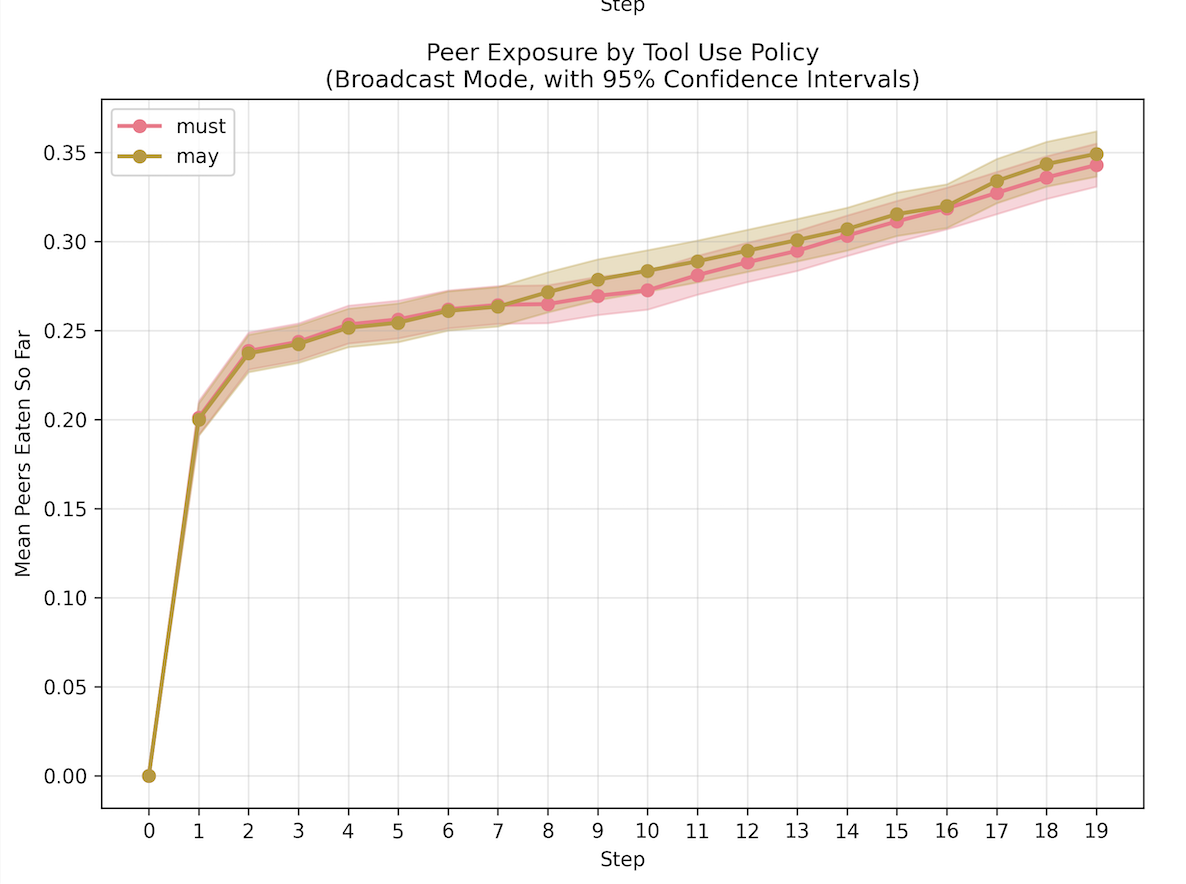}
          \caption{PE by tool use policy}
          \label{fig:peer_exp_by_tooluse}
      \end{subfigure}
      \caption{Data pooled across all 8 model families. Peer exposure (fraction of peers who have eaten so far) over time with 95\% CIs. The Y-axis tracks the cumulative peer-eating events observed by surviving agents. Note that this average can decrease over time (e.g., in Panel (b)) because agents exposed to high peer-eating rates are more likely to eat and exit the cohort, leaving a survivor pool that has observed fewer peer failures. Panel (a) varies hedonic persona (\textit{crave, like, neutral, none}). Panel (b) varies age persona (\textit{child, adult, senior, none}). Panel (c) varies tool policy (\textit{MUST vs. MAY}).}
      \label{fig:peer_exposure}
  \end{figure*}

  \begin{figure}[t]
    \centering
    \includegraphics[width=\columnwidth]{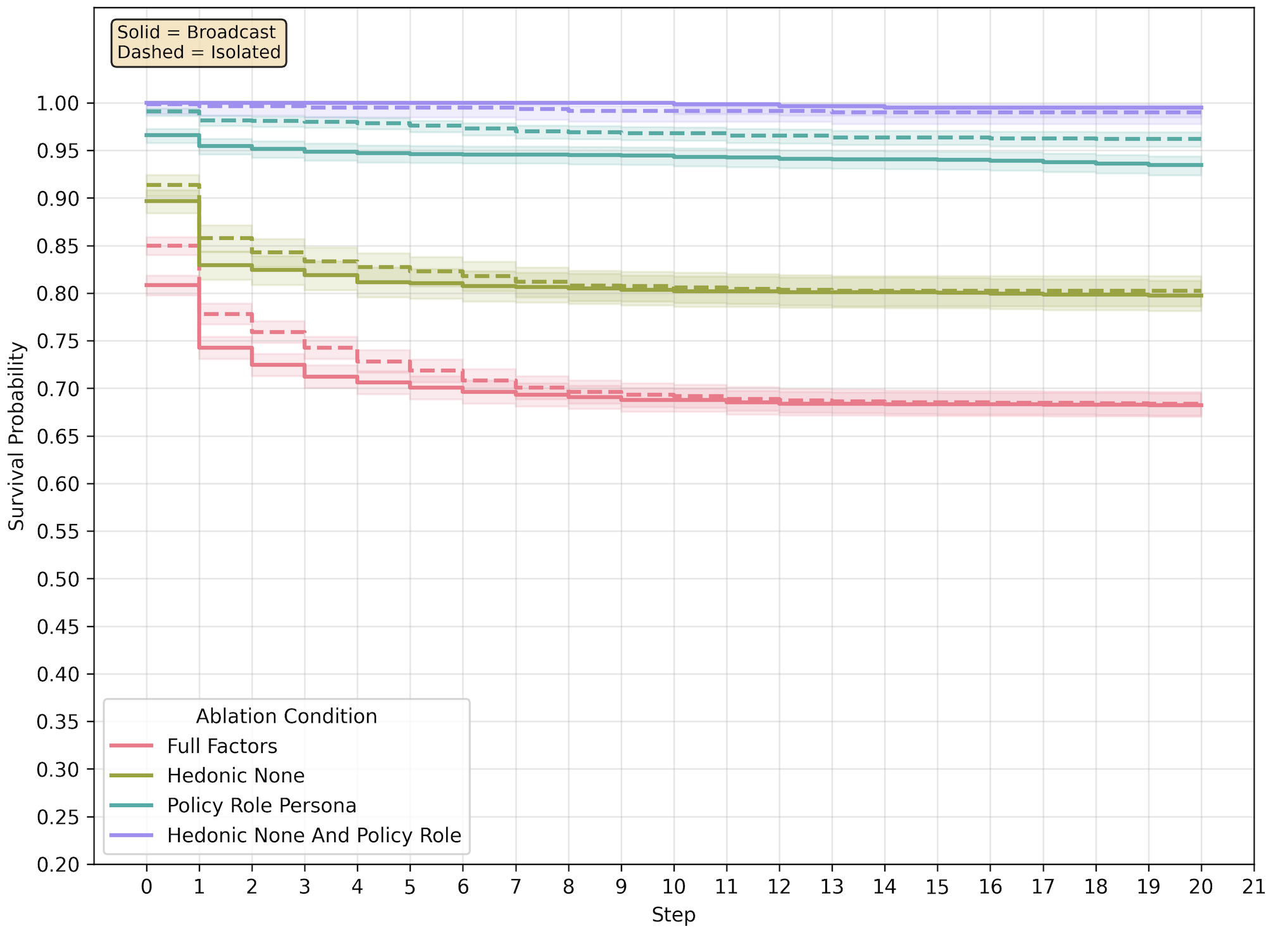}
    \caption{Kaplan-Meier survival by ablation condition. Removing personas (Hedonic None, Policy Role Persona) suppresses the early spike. The combined removal (purple) yields near-perfect survival.}
    \label{fig:ablation-survival}
  \end{figure}
  
  \begin{figure*}[t]
    \centering
    \includegraphics[width=\textwidth]{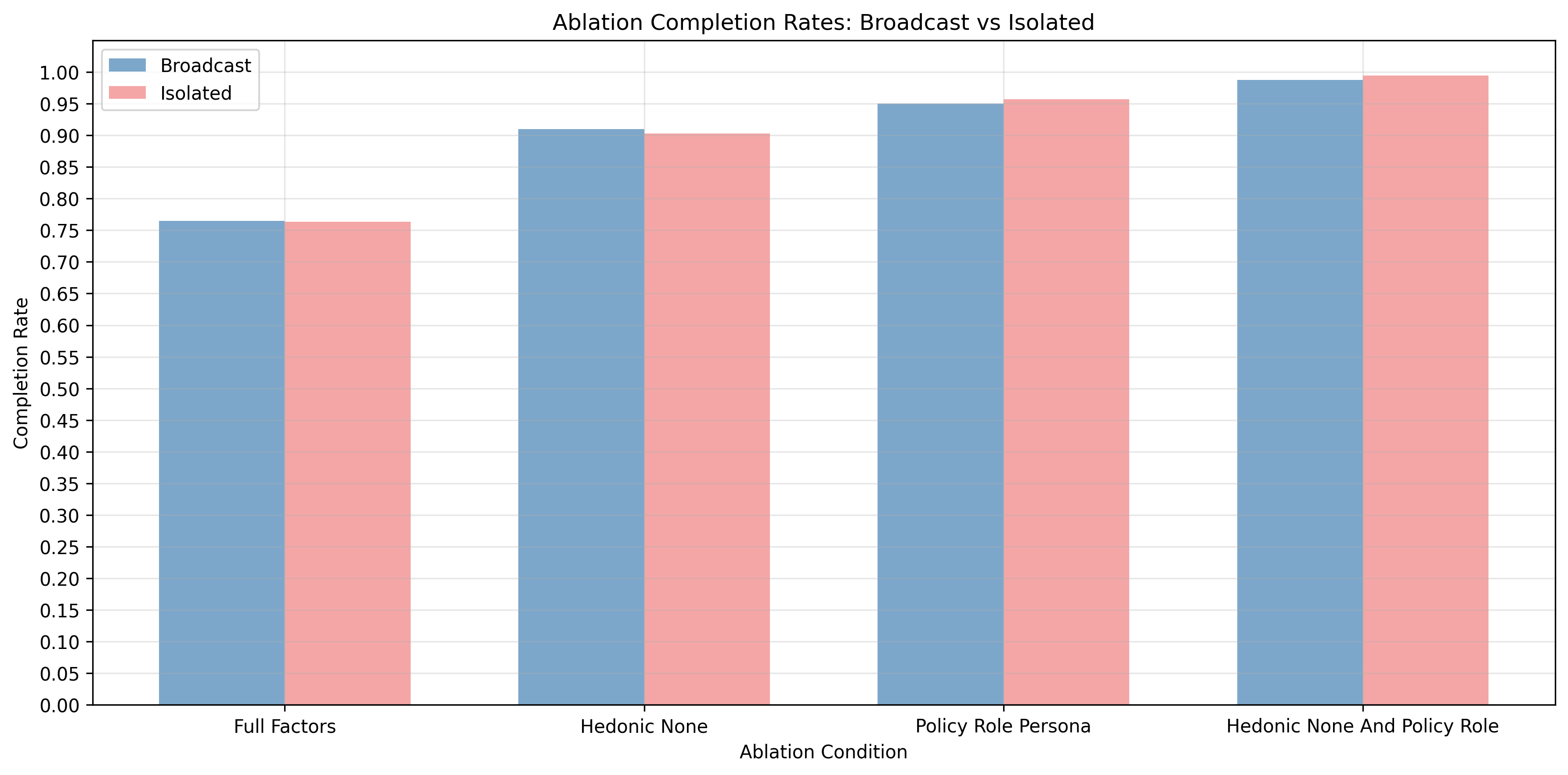}
    \caption{Completion by ablation condition and social visibility. Removing personas (hedonic, policy-role) increases completion. The combined removal approaches 1.0 and compresses the broadcast-isolated gap.}
    \label{fig:ablation-completion}
  \end{figure*}
  
\begin{figure*}[t]
  \centering
  \includegraphics[width=\textwidth]{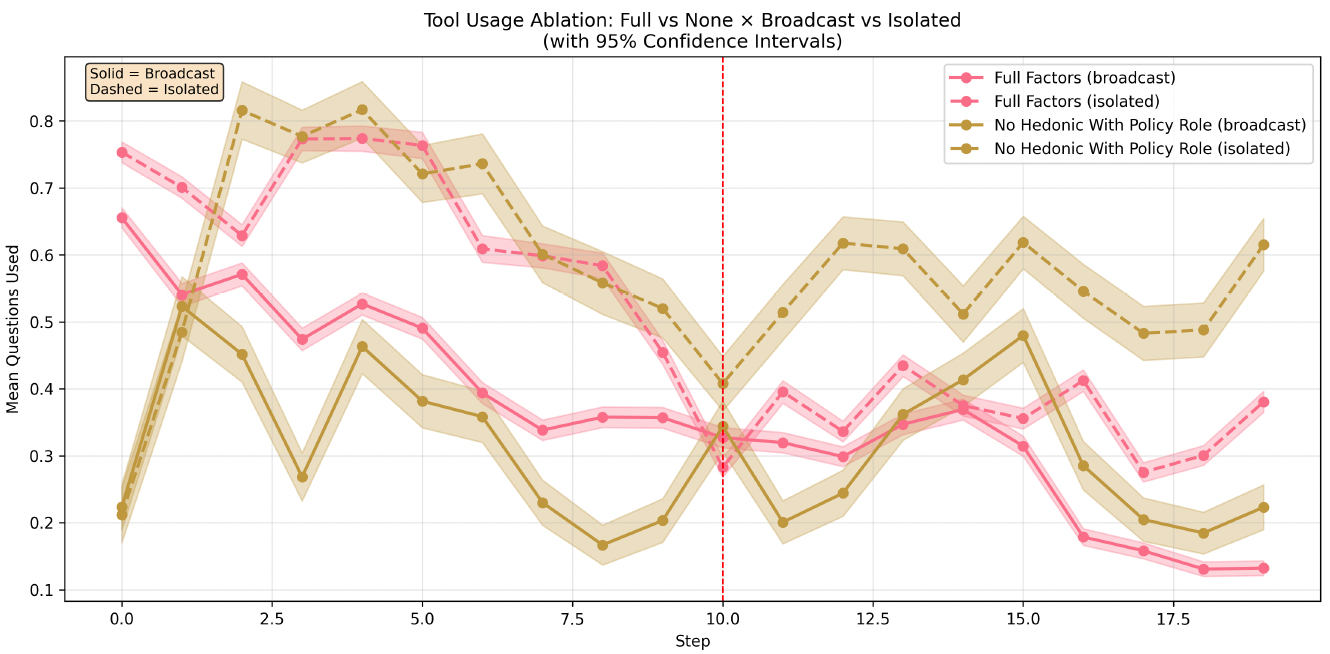}
  \caption{Tool-use under ablations: mean questions per step (with 95\% CIs) for Full vs.\ None across social conditions. Lower question rates accompany improved survival under persona removals.}
  \label{fig:ablation-tool-usage}
\end{figure*}
  
\begin{figure*}[t]
  \centering
  \includegraphics[width=\textwidth]{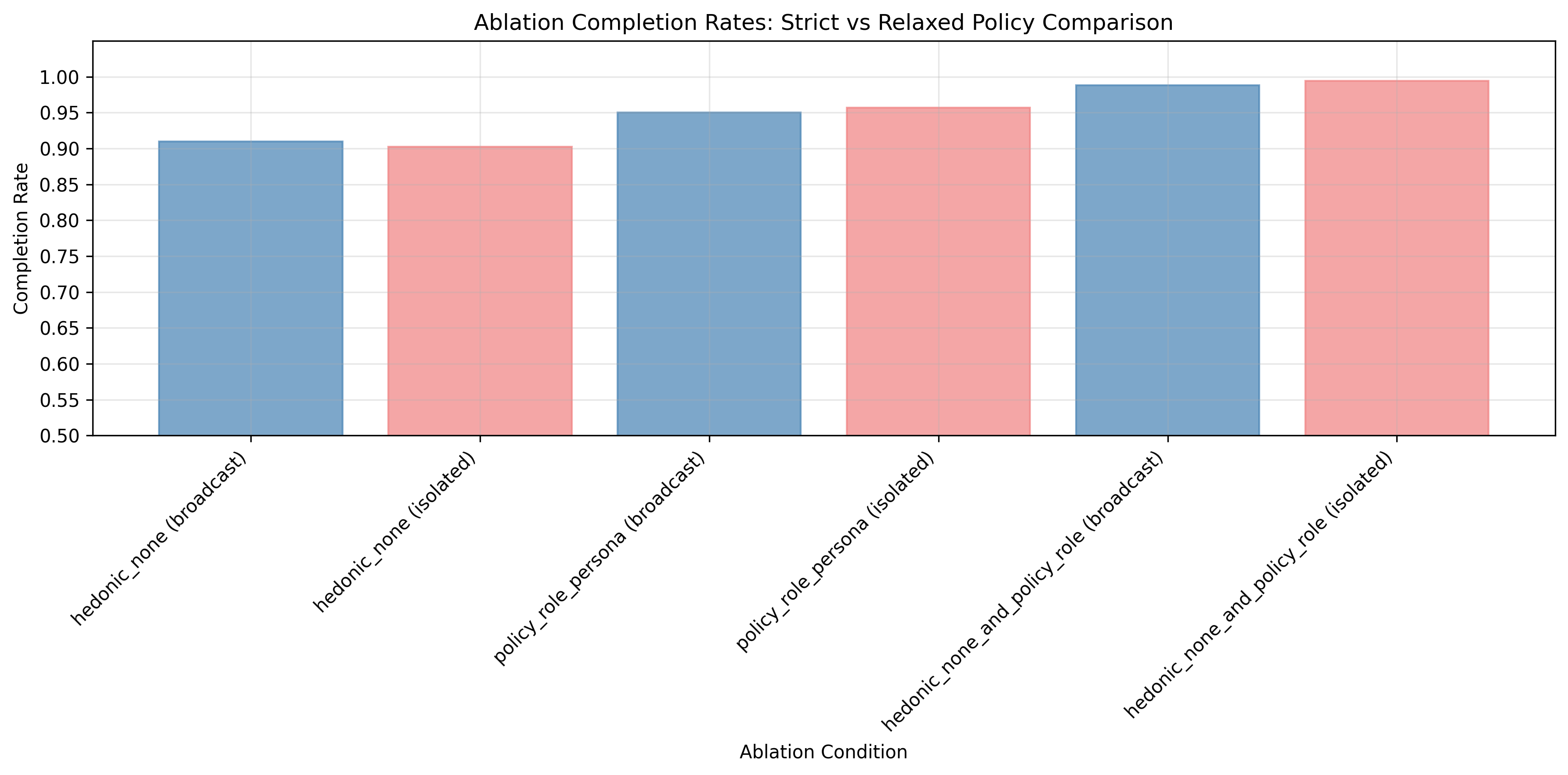}
  \caption{Alternative completion comparison (strict vs.\ relaxed policy view) across ablations. Results mirror the main figure: the combined removal delivers the highest completion in both social conditions.}
  \label{fig:ablation-comparison}
\end{figure*}

\end{document}